\documentclass{article}
\usepackage[utf8]{inputenc}
\usepackage[margin = 1in]{geometry}
\usepackage{authblk}
\usepackage{subfigure}

\usepackage{amsmath, amssymb, amsthm, amsfonts, mathtools, multirow, mathrsfs, tabu, booktabs, lipsum, url, graphicx, xcolor, tabu,  authblk, hyperref, blindtext}

\usepackage[english]{babel}
\usepackage[shortlabels]{enumitem}
 \usepackage[ruled,vlined, linesnumbered]{algorithm2e}
\usepackage[capbesideposition=outside,capbesidesep=quad]{floatrow}
\usepackage[framemethod=tikz]{mdframed}
\usepackage{bbm}

\allowdisplaybreaks

\newcommand{\appropto}{\mathrel{\vcenter{
  \offinterlineskip\halign{\hfil$##$\cr
    \propto\cr\noalign{\kern2pt}\sim\cr\noalign{\kern-2pt}}}}}
    
\usepackage[backend=biber,style=ieee,sorting=none,natbib=true]{biblatex} 

\usepackage[utf8]{inputenc}
\usepackage[autostyle=true, threshold=2]{csquotes}

\def\vc{{\bf c}}
\def\vd{{\bf d}}

\def\vr{{\bf r}}

\def\vx{{\bf x}}
\def\vy{{\bf y}}

\title{\textbf{A 3-stage Spectral-spatial Method for Hyperspectral Image Classification}}

\author[1,2]{Raymond H. Chan}
\author[1]{Ruoning Li\footnote{Corresponding Author: ruoningli3-c@my.cityu.edu.hk}}

\affil[1]{ Department of Mathematics, City University of Hong Kong, Hong Kong}
\affil[2]{ Hong Kong Centre for Cerebro-Cardiovascular Health Engineering, Hong Kong}

\date{}                     
\begin{document}
\topmargin=0mm

\maketitle
\begin{abstract}
Hyperspectral images often have hundreds of spectral bands of different wavelengths captured by aircraft or satellites that record land coverage. Identifying detailed classes of pixels becomes feasible due to the enhancement in spectral and spatial resolution of hyperspectral images.
In this work, we propose a novel framework that utilizes both spatial and spectral information for classifying pixels in hyperspectral images. The method consists of three stages. In the first stage, the pre-processing stage, Nested Sliding Window algorithm is used to reconstruct the original data by {enhancing the consistency of neighboring pixels} and then Principal Component Analysis is used to reduce the dimension of data.  In the second stage, Support Vector Machines are trained to estimate the pixel-wise probability map of each class using the spectral information from the images. Finally, a smoothed total variation model is applied to smooth the class probability vectors by {ensuring spatial connectivity} in the images.
We demonstrate the superiority of our method against three state-of-the-art algorithms on six benchmark hyperspectral data sets with 10 to 50 training labels for each class. The results show that our method gives the overall best performance in accuracy. Especially, our gain in accuracy increases when the number of labeled pixels decreases and therefore our method is more advantageous to be applied to problems with small training set. Hence it is of great practical significance since expert annotations are often expensive and difficult to collect.
\end{abstract}

\noindent \textbf{Index Terms}: 
hyperspectral classification; semi-supervised learning; nested sliding window;  support\\ vector machines; smoothed total variation; image reconstruction.

\section{Introduction}
{Hyperspectral images} (HSIs) often have hundreds of electromagnetic bands of reflectance collected by {aircraft} or satellites that contain rich spectral and spatial information. They can be represented by a tensor ${\cal X} \in \mathbb{R}^{M\times N\times B}$, where $M, N$ are the rows and columns of the image in each spectral band and $B$ is the number of bands~\cite{Eismann2012HRS}. In general, each distinct material has its own spectral signature in HSIs owing to its unique chemical composition.
The improvement in spectral resolution makes it possible to explore the HSIs using machine learning approaches in various applications like land coverage mapping, change recognition, water quality monitoring, and mineral identification \cite{Morchhale2016dCNN, Liu2019review, Khan2018modern, Peyghambari2021lithological}. The rich information in HSIs enables the algorithms to distinguish more detailed categories for land cover clustering and classification so that HSIs play a vital role in detecting different natural resources and monitoring vegetation health \cite{Cui2021ANG, IM2012Vegetation, Horig2001HyMap, Qin2016Oil, Jin2018Wheat, DVIS, Madingley, ADVIS, Neupane2021Plant, Camalan2022,Aland}.

A variety of algorithms with and without manual annotations have been developed for the classification of pixels in HSIs. Compared with unsupervised methods, semi-supervised methods require a few labeled data for training, and they can produce considerable improvement in performance.
The classical pixel-wise semi-supervised algorithms, such as support vector machines (SVMs) \cite{Melgani2004SVM}, k-nearest-neighbor (kNN) classifier \cite{Kuo2008knn}, multinomial logistic regression \cite{Li2013Soft}, and random forest \cite{Ham2005RF, Breiman2001RF}, were applied in the past.

These classifiers only explore and analyze the spectral information of HSIs, whereas the spatial information is not utilized. For regions that are spatially homogeneous but with a {variety} in the spectra, these methods may produce a noisy classification map (see e.g., Figure \ref{IndianPines_result}(d)). A common theoretical assumption in HSI classification is local spatial connectivity in certain regions, which means spatially nearby pixels have a higher probability belonging to the same class \cite{Bo2017Weighted, Liu2013KSR}. Thus, pixel-wise classification methods can be enhanced by incorporating the spatial dependency of the pixels. Basically, the utilization of spatial information can take place in the pre-processing and post-processing steps of HSI classification \cite{Ren2021NSW, Li2022SaR, Chan2020twostage}.

Chan et al. \cite{Chan2020twostage} incorporate segmentation techniques in their 2-stage method to incorporate spatial information as a post-processing step. After acquiring the class probability vector for each pixel by SVM, a convex variant of the Mumford-Shah method (equivalent to a smoothed total-variational method) is used to denoise the probability vectors. Experiments show that this method improves the accuracy significantly, see Figure \ref{IndianPines_result}{(f)}. The authors in \cite{Camps-Valls2006Composite} redefine a pixel in both spectral domain and spatial domain by extracting features in its neighboring region. Then Mercer's kernels are adopted in SVM to combine spectral and spatial information. Structural filtering methods, for instance, the Gabor filter, can extract spatial texture features of adjacent pixels in different scales and directions \cite{Rajadell2013Gabor, Bau20103D}.
Mathematical morphology can be used to obtain the morphological profile, such as the orientation or size of the spatial structures of images \cite{Fauvel2013Advances}.
Fang et al. \cite{Fang2017MFASR} propose an adaptive sparse representation (MFASR) method based on four {spatial and spectral} features where spatial information is extracted by the Gabor filter, extended morphological profiles and differential morphological profiles.
Gan et al. \cite{Gan2018MFKSR} propose a multiple feature kernel sparse representation-based classifier, which transforms each feature into a low-dimensional space with a nonlinear kernel.

In the extreme sparse multinomial logistic regression framework, the extended multi-attribute profile is adopted for spatial feature extraction \cite{Cao2017ESMLR}.
Gao et al. \cite{Gao2018RMG} propose a new approach that extracts spatial feature by applying linear prediction error and the local binary pattern. It then combines the spatial and spectral information by using a vector stacking method before feeding into the {Random Multi-Graphs model, which is proposed in \cite{Zhang2017RMG}.}  The $K$-means algorithm and principal component analysis (PCA) are adopted in \cite{Shu2018Learning} to extract spatial features, and then a SVM is trained to produce the classification results. Ren et al. \cite{Ren2021NSW} propose the Nested Sliding Window (NSW) pre-processing method to extract spatial information from original HSI data. The NSW algorithm determines the optimal sub-window position based on the average Pearson correlation coefficient of the target pixel and its neighboring pixels, and then the pixels are reconstructed depending on the pixels in the sub-window and their correlation coefficients. PCA is used to further process the reconstructed data for dimensionality reduction and denoising. Finally the reconstructed data are fed into SVM for classification.

The convolutional neural network (CNN) is becoming popular these years which can extract spatial information internally by convolutional kernels. The original CNNs {\cite{OCNN}} learn spatial features naturally from the original images by generating some convolutional layers. Gao et al. \cite{Gao2018CNN} employ a new CNN architecture that also takes the spatial features extracted from the original image as input and achieves a significant improvement of accuracy compared with the original CNN framework. Zhang et al. \cite{Zhang2018DRCNN} create a diverse region-based CNN which learns spatial features based on inputs from different regions. The recurrent 2-D CNN and recurrent 3-D CNN achieve higher accuracies and faster convergence rates {with} its convolutional operators and the recurrent network structure \cite{Yang2018DL}.
Nonetheless, these CNNs have millions of parameters that need to be tuned. Thus they require powerful machines to train the model and a large number of expert labels that are expensive to get. It is more feasible and preferable to only incorporate a few labeled pixels for training as in the semi-supervised learning methods \cite{Ren2021NSW}.

In this work, we propose a 3-stage method for HSI classification. The first stage is a pre-processing step where we first apply the NSW algorithm \cite{Ren2021NSW} to find the most correlated nested window and then reconstruct the data based on the Pearson correlation for each pixel. Then we use PCA to reduce the dimension of the reconstructed data.  In the second stage,  we train {an SVM-type method $\nu$SVC ($\nu$-support vector classifier) \cite{Scholkopf2000SVM}} for semi-supervised classification and produce an estimated probability tensor consisting of the probability maps for all classes. In the last stage, to incorporate the spatial information, a smoothed total variation  model \cite{Chan2020twostage} is applied to post-process the probability maps to remove isolated misclassified pixels.

{To demonstrate the {efficacy} of our method, we test it against three state-of-the-art methods on six widely used benchmark hyperspectral data sets with 10 to 50 training labels for each class.}
{The results show that our method gives the best overall accuracy on all six data sets except on the PaviaU data set when 10 pixels in each class are given. {Even in that case}, our method achieves the second highest accuracy. We emphasize that our gain in accuracy is higher when the number of labeled pixels is smaller. Our method is therefore of great practical significance since expert annotations are often expensive and difficult to collect.}

{The superiority of our method stems {from} the fact that the spatial information of the image is fully explored. The pre-processing step enhances the {consistency} of adjacent pixels, especially for those pixels which are located in a large homogeneous area and have similar interclass spectra. }
{Through the reconstruction, the similarity of spectral information of the pixels in the same category can be utilized so that we need a smaller set of training pixels for each class.}
{This step is useful for data sets which do not have sufficiently good spectral information. The post-processing step further improves the classification result by ensuring the smoothness across spatial homogeneous regions using the spatial positions of the training pixels. The smoothed total variation model used here can simultaneously enhance the spatial homogeneity by denoising while segmenting the image into different classes.}

This paper is organized as follows. Section 2 introduces the three stages of our method. Sections 3 and 4 give the numerical results and discussions on six benchmark data sets. Section 5 concludes the experiments and discusses the planned future work.

\section{The proposed method}\label{Pre}
\begin{figure}[H]
\includegraphics[width=\textwidth]{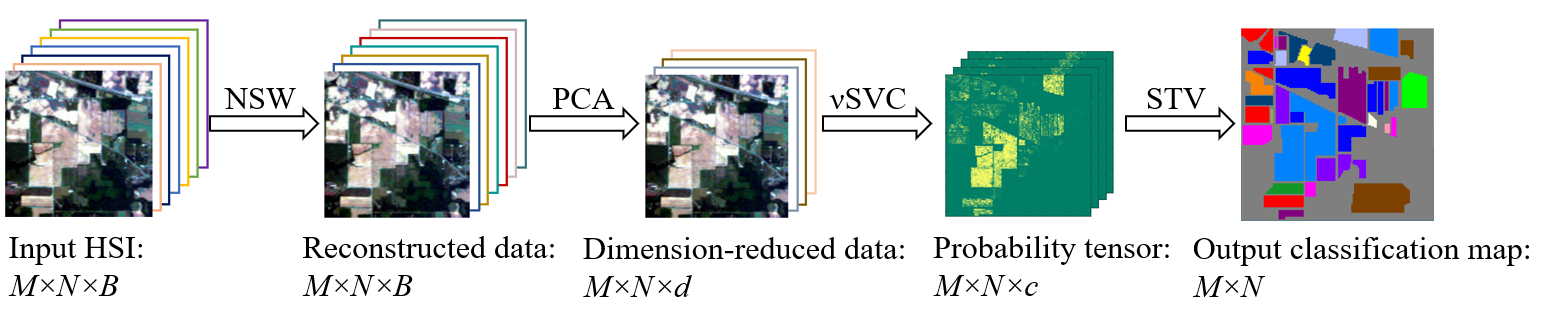}
\caption{The outline of the proposed method, where $d$ is the reduced dimension and $c$ is the number of classes.
\label{outline1}}
\end{figure}

The method proposed in this work comprises of the following three stages:
(i) pre-processing stage: the HSI data set is reconstructed by NSW and then projected linearly to a lower-dimensional space by PCA. The step effectively use spatial information and reduce the Gaussian white noise in HSIs \cite{Ren2021NSW, Luo2016PCA};
(ii) pixel-wise classification stage: the $\nu$SVC, which uses mainly the spectral information in the data set, is applied to get the probability maps where each map gives the probability of the pixels belonging to a certain class \cite{Chan2020twostage, Scholkopf2000SVM, Hsu2002OAO, Lin2007Probability, Wu2004Probability};
(iii) smoothing stage: a smoothed total variation (STV) model is used to {ensure local spatial connectivity} in the probability maps so as to increase the classification accuracy {\cite{ Chan2020twostage,Mumford1989OptimalAB}}. In the following subsections, we introduce the three stages in detail. The outline of the whole method is illustrated in Figure~\ref{outline1}.

\subsection{The pre-processing stage}

The pre-processing of the HSI data sets can effectively improve the quality of the data, leading to a better performance in classification
{with less number of training pixels} \cite{ Ren2021NSW}. In the pre-processing, spatial features are usually extracted by analyzing the similarity between the spectral values of the pixels in local regions. Wu et al. \cite{Fu2016SAJSR} generate a shape-adaptive region for each target pixel by applying LPA-ICI method, and then put them together into the joint sparse representation classifier, which effectively explores the spatial information. On this basis, in \cite{Li2022SaR}, a shape-adaptive reconstruction method is proposed to reconstruct data based on the shape-adaptive region. Bazine et al. \cite{Bazine} propose a CDCT-WF-SVM model where the original data are pre-processed by applying spectral DCT and spatial filtering adaptive Wiener filter to extract the most significant information before SVM. {The {NSW} method in \cite{Ren2021NSW} is to find the best nested sliding window at each pixel with the largest mean Pearson correlation coefficient and then reconstruct the given pixel by weighting pixels using normalized correlation coefficients in the best window.} {It then uses PCA to reduce the dimension of the reconstructed data. We adopt this approach in our pre-processing stage and explain it briefly in the following two subsections, { and we refer the details to \cite{Ren2021NSW}}.}

\subsubsection{The Nested Sliding Window (NSW) Method}
For two pixels $\vx$ and $\vy$, the Pearson correlation coefficient is defined as:
\begin{equation}
  \begin{split}
  {\rm corr}(\mathbf{x},\vy)=\frac{{\rm Cov}(\vx,\vy)}{\sqrt{{\rm Var}(\vx)\cdot {\rm Var}(\vy)}},\label{eq1}
  \end{split}
\end{equation}
where ${\rm Cov}(\vx,\vy)$ represents the covariance between $\vx$ and $\vy$, and ${\rm Var}(\cdot)$ is the variance.
We define the neighboring pixels of a target pixel $\vx_{ij}$ with a window size $\omega$ as
$$
  {\cal N}({\rm\vx}_{ij})=\{{\rm \vx}_{mn}\mid m\in [i-a,i+a], n\in [j-a,j+a]\},
$$
where $a+1\leq i\leq M-a, a+1\leq j\leq N-a$, and $ a=(\omega -1)/2$.  {For {target} pixels on or near the boundary of the image, we use zero-padding to extend the image outside the boundary so as to obtain a window of the same size $\omega$ for these pixels.}

{Then we create a series of sliding windows inside ${\cal N}$ to search for some neighboring pixels which are most similar to the target pixel  $\vx_{ij}$.} In order to calculate the correlation coefficients between the target pixel and its neighboring pixels, each sliding window should contain the target pixel, that is, the size of the sliding window should be (($a+1$),($a+1$)). Then the neighboring pixels with a sliding window can be expressed as the 3-D tensor:
$$
  {\cal S}_{pq}({\vx}_{ij})=\{{\rm \vx}_{mn}\mid m\in [i-a+p,i+p], n\in [j-a+q,j+q]\} {\in \mathbb{R}^{(a+1) \times (a+1) \times B}},
$$
where $a+1\leq i\leq M+a$ and $a+1\leq j\leq N+a$. Here $0<p,q\leq a$ determine the position of the sliding window.

Thus, the Pearson correlation coefficient between the target pixel and each neighboring pixel in the sliding window can be computed by (\ref{eq1}). Together, they form a matrix in each sliding window, denote as $C_{pq}({\rm\vx}_{ij})$:
$$
  C_{pq}({\rm\vx}_{ij})=\{c_{mn}|m\in [i-a+p,i+p], n\in [j-a+q,j+q]\}.
$$
{Let $\overline{c}_{pq}$ be the mean value of all the entries in  $C_{pq}$.} After going through all the sliding windows,  we set $c_{kl}\equiv \max \overline{c}_{pq}$. 

The corresponding 3D tensor ${\cal S}_{kl}(\vx_{ij})$ and the 2D matrix $C_{kl}(\vx_{ij})$ are re-shaped to a 2D matrix ${S}_{kl}(\vx_{ij})$ and a vector $\vc_{kl}(\vx_{ij})$ with size $((a+1)\times(a+1),B)$ and size $((a+1)\times(a+1),1)$, respectively. The correlation coefficient vector $\vc_{kl}(\vx_{ij})$ is then rescaled by
$$
   \vc_{kl}(\vx_{ij})=\vc_{kl}(\vx_{ij})/{\rm Sum}(\vc_{kl}(\vx_{ij})).
$$
The reconstruction of the pixel $\vx_{ij}$ at the $(i,j)$ location is given by the $B$-vector:
$$
  {\vr_{ij}={S}_{kl}(\vx_{ij})^{\top}\vc_{kl}(\vx_{ij})}.
$$
{After the reconstruction processing, we obtain an $R\in \mathbb{R}^{B \times (MN)}$ whose columns are $\vr_{ij}$.}

\subsubsection{Principal Component Analysis (PCA)}
PCA \cite{Goodfellow2016DL} is one of the most commonly used dimensionality reduction algorithms. {Assume that we need to {reduce} the data $R$ obtained by NSW algorithm from $B$ dimensional to $d$ dimensional}, then the purpose of PCA is to find a 2D transformation matrix $W \in \mathbb{R}^{B \times d}$ in
$\arg\max_{W}{\rm tr}(W^{\top}RR^{\top}W)$
where ${\rm tr}(\cdot)$ represents the trace of the matrix and  $W^{\top}W=I$. The maximization of $W$ can be solved by using the Lagrangian multiplier method. Finally, we get the dimension-reduced data
\begin{equation}\label{D}
D=W^{\top}R \in \mathbb{R}^{d \times (MN)}.
\end{equation}

\subsection{The pixel-wise classification stage}
Support vector machines (SVMs) have been used successfully in pattern recognition \cite{Pontil1998SVM}, object detection \cite{EI2002SVM,Osuna1997SVM}, and financial time series forecasting \cite{Tay2001SVM,Kim2003SVM} etc., to separate two classes of objects. {Suppose we have $t$ labeled pixels, then} the formulation of $\nu$SVC is given as follows:
\begin{equation}
    \left\{
        \begin{array}{ll}
            \min \limits_{\mathbf{w},b,\boldsymbol{\xi},\rho} \frac{1}{2}\|\mathbf{w}\|_2^2-\nu \rho+\frac{1}{t}\sum\limits_{i=1}^t\xi_i\\
            {\rm s.t.}\\
            y_i(\mathbf{w}^{{\top}} \phi (\vd_i)+b)\geq \rho -\xi_i,\ i=1,2,...,t,\\
            \xi_i\geq 0,\ i=1,2,...,t,\\
            \rho \geq 0,\label{eq8}
        \end{array}
    \right.
\end{equation}

where $\vd_i \in \mathbb{R}^{d}$ is the column in the matrix $D$ in (\ref{D}), $i$ represents the $i$-th labeled pixel, $\phi$ is a feature map that maps the data to a higher dimensional space in order to improve the separability between the two classes, $\mathbf{w}$ and $b$ are the normal vector and the bias of the hyperplane respectively, and $\xi_i$ is the slack variable.

Model (\ref{eq8}) can be solved by its Lagrangian dual. Finally we obtain the hyperplane function $\mathbf{w}^{{\top}} \phi(\vd)+b$ which is then used to classify each test pixel $\vd\in \mathbb{R}^{d}$ (which are columns of $D$ in (\ref{D})), see \cite{Cortes1995SVM}. In the experiments, we follow  \cite{Cortes1995SVM} and use radial basis functions  for $\phi(\cdot)$ where its parameter is determined by a 5-fold validation.
{Under the {one-against-one} strategy, there are $[c(c-1)]/2$ such pairwise hyperplane functions where $c$ is the number of classes. We use them to estimate the probability $p_k$ that a non-labeled pixel $\vd$ is in class $k$, see \cite{Lin2007Probability, Wu2004Probability}.} {Finally,  we obtain a 3D tensor ${\cal V} \in \mathbb{R}^{M \times N \times c}$ where ${\cal V}_{i,j,k}$ denotes the probability that the pixel $\vd$ at the $(i,j)$ location is in class $k$, and ${\cal V}_{:,:,k}$ denotes the probability map for class $k$.}
{In particular, if a pixel $(i, j)$ is a training pixel belonging to the $l$-th class, then ${\cal V}_{i,j,l} = 1$ while ${\cal V}_{i,j,k} = 0$ for all other $k$’s.}

\subsection{The smoothing stage}
Post-processing the probability maps can further improve the performance.
Markov Random Field regularization is applied to post-process the classification results by considering spatial and edge information in \cite{Tarabalka2010SVM}. In \cite{Chakravarty2017Fuzzy}, Fuzzy-Markov Random Field is adopted to smooth the classification result predicted by SVM. In our previous work \cite{Chan2020twostage, Li2022SaR, Chan2013seg, Chan2017seg}, a smoothed total variation (STV) model is proposed to denoise the probability maps ${\cal V}_{:,:,k}$ that $\nu$SVC produces by {ensuring local spatial connectivity} in the maps. We adopt the same model here in our method.

Let $V_k={\cal V}_{:,:,k},k=1,...,c$. In this stage, we enforce the local connectivity by minimizing:
\begin{equation}\label{eq4}
    \left\{
        \begin{array}{ll}
         \min\limits_{{U}_k} \frac{1}{2}\| {U}_k-{V}_k\|_2^2+\beta_1\|{{\nabla U}}_k\|_1+\frac{\beta_2}{2}\|{{\nabla U}}_k\|_2^2\\
         {\rm s.t.}\ {U}_k|_\Omega={V}_k|_\Omega
        \end{array}
    \right.
\end{equation}
where $\beta_1$ and $\beta_2$ are the regularization parameters and $\Omega$ denotes the training set {so that the constraint can keep the classifications of the training pixels unchanged.} {The operator $\nabla$ means the discrete gradient of the matrix $U_k$ when considering it as a 2-D image.}

This is an $\ell_1$-$\ell_2$ problem and can be solved by the alternating direction method of multipliers (ADMM) \cite{Boyd2011ADMM}. The minimizer $U_k$ is the enhanced probability map for class $k$. When the probability map for each class  is obtained, we get a 3D tensor ${\cal U}$ {where ${\cal U}_{:,:,k}=U_k$}. {The final classification for the pixel $(i,j)$ is then given by $\mathop{\arg\max}\limits_{k\in {\{1,...,c\}}}{\cal U}_{i,j,k}$.}

\section{Experimental Results}
In this section, we compare our method with {the classical} $\nu$SVC and three other state-of-the-art methods on six commonly used data sets.

\subsection{Data Sets}
In order to test the superiority of our method, six widely used publicly available hyperspectral data sets are chosen for testing. They are the Indian Pines, Salinas, Pavia Center, Kennedy Space Center (KSC), Botswana, and University of Pavia (PaviaU) data sets. They have different sizes and different number of spectral bands of different wavelengths, and they are commonly used these years in the study of hyperspectral images. In the following we introduce them one by one.

The Indian Pines data set was collected in the test site located in the Northwest India by the AVIRIS sensor. It consists of 145$\times$145 pixels and each pixel has 220 spectral reflectance bands with the wavelength from 0.4 to 2.5 $\mu$m. After eliminating the effect of water absorption, the number of bands finally is 200. Its ground-truth consists of 16 classes.

The Salinas data set was collected over Salinas Valley in California by the AVIRIS sensor with high spatial resolution of 3.7m per pixel. The size is 512$\times$217 {pixels} with 224 spectral reflectance bands. Same as the Indian Pines data set, due to the water absorption, the number of band decreases to 204 after discarding the 108th--112th, 154th--167th, and 224th bands. There are 16 classes in Salinas data set.

The Pavia Center data set and PaviaU data set were acquired by the ROSIS sensor over Pavia in Italy with spatial resolution of 1.3m. The data set sizes are 1096$\times$715$\times$102 and 610$\times$340$\times$103 respectively, where 102 and 103 represent the numbers of the spectral bands respectively. Both data sets have 9 classes.

The KSC data set was acquired over the Kennedy Space Center in Florida by the NASA AVIRIS sensor. It has 224 bands with wavelength from 0.4 to 2.5 $\mu$m but after removing water absorption and low SNR bands, it has 176 bands totally. The size is 512$\times$614 pixels and there are 13 classes. The sensor has a spatial resolution of 18m.

The Botswana data set was collected by the Hyperion sensor on NASA EO-1 satellite over Botswana with 30m resolution. It has 145 bands after removing 97 bands because of water absorption and covers the wavelength from 0.4 to 2.5 $\mu$m. The area is of size 1476$\times$256, and there are 14 classes in the ground-truth.

\subsection{Comparison Methods and Evaluation {Metrics}}

We compare our new method with several currently used methods: $\nu$-support vector classifier ({$\nu$SVC}) \cite{Scholkopf2000SVM}, multiple-feature-based adaptive sparse representation (MFASR) \cite{Fang2017MFASR}, the 2-stage method \cite{Chan2020twostage}, and NSW-PCA-SVM \cite{Ren2021NSW}. We remark that in \cite{Chan2020twostage,Ren2021NSW} there are comprehensive comparisons of the last two methods with many other methods which show the superiority of these two methods with others.

In this paper, we use Overall Accuracy (OA), Average Accuracy (AA) and kappa coefficient (kappa) to quantitatively evaluate the performance of these five methods \cite{metrics}. For each method, ten runs were conducted. In order to ensure the reliability of the experiments, the training set was randomly selected for each run and finally the average of the results obtained from the ten runs was taken for comparison. In each figure, there is an error bar (the color bar) which represents the number of misclassification for each {pixel} in the image over the ten runs. {As in \cite{Ren2021NSW} and \cite{Fang2017MFASR}, we assume the background pixels are given and we do not classify them. We only compare the accuracies on the non-background pixels.}

All the tests were run on a computer with an Intel Core i7-9700 CPU, 32 GB RAM and the software are MATLAB R2021b and Python 3.9.

\subsection{Classification Results}
Table~\ref{IndianPines} shows the {average} classification results of each method for the Indian Pines data set, which has large homogeneous regions with more regular shapes. In each experiment, 10 training pixels for each class were randomly selected and the remaining pixels were used for testing. {The table shows the average accuracy over 10 {runs}}, and we use boldface font to denote the best results among the methods. We see that our method generates the best results for all three metrics (OA, AA and kappa) and is at least 2.61\% higher than the results of all other methods. For some classes with a small number of pixels, like Alfalfa, Grass/pasture-mowed and Oats, the results of our method achieve the highest accuracy, reaching 100\%. For classes like Corn-no till and Soybeans-mill till with higher misclassification rate under the 2-stage method, the rates are enhanced a lot in our method. This illustrates the power of the the pre-processing stage in our method.

\begin{table}[htb]
\centering
\caption{{Average classification accuracies over 10 trials} for the Indian Pines data set with 10 random training pixels for each class.}
\label{IndianPines}
\resizebox{0.8\textwidth}{!}{%
\begin{tabular}{cccccc}
\hline
\textbf{Class}	& \textbf{$\nu$SVC}	& \textbf{MFASR} & \textbf{2-stage method} & \textbf{NSW-PCA-SVM} & \textbf{Our method}\\
\midrule
Alfalfa	& 82.22\%&97.50\%&98.89\%&97.50\%&\textbf{100\%}\\
Corn-no till & 39.32\%&71.09\%&75.05\%&77.35\%&\textbf{83.50\%}\\
Corn-mill till &49.05\%&80.09\%&91.26\%&89.68\%&\textbf{92.29\%}\\
Corn &63.83\%&88.28\%&\textbf{100\%}&89.74\%&97.22\%\\
Grass/pasture &77.61\%&84.93\%&88.37\%&86.17\%&\textbf{92.66\%}\\
Grass/trees	&80.97\%&92.61\%&99.04\%&97.44\%&\textbf{99.82\%}\\
Grass/pasture-mowed	&93.33\%&\textbf{100\%}&\textbf{100\%}&\textbf{100\%}&\textbf{100\%}\\
Hay-windrowed &72.12\%&99.76\%&\textbf{100\%}&99.83\%&\textbf{100\%}\\
Oats &96.00\%&\textbf{100\%}&\textbf{100\%}&\textbf{100\%}&\textbf{100\%}\\
Soybeans-no till  &52.92\%&81.94\%&85.21\%&87.43\%&\textbf{89.59\%}\\
Soybeans-mill till &42.76\%&68.81\%&66.72\%&78.00\%&\textbf{86.01\%}\\
Soybeans-clean &36.59\%&83.36\%&\textbf{90.81\%}&80.57\%&90.63\%\\
Wheat &92.36\%&99.49\%&99.59\%&98.67\%&\textbf{99.95\%}\\
Woods &67.55\%&91.51\%&94.96\%&95.24\%&\textbf{99.58\%}\\
Bridg-Grass-Tree-Drives	&41.81\%&94.95\%&97.23\%&96.54\%&\textbf{97.61\%}\\
Stone-steel lowers &93.61\%&98.92\%&99.88\%&97.23\%&\textbf{100\%}\\ \hline
OA	&54.31\%&81.30\%&84.42\%&86.48\%&\textbf{91.57\%}\\
AA	&67.63\%&89.58\%&92.94\%&91.96\%&\textbf{95.55\%}\\
kappa &49.00\%&78.90\%&82.54\%&84.68\%&\textbf{90.42\%}\\
\hline
\end{tabular}%
}
\end{table}

\begin{figure}[ht]
\includegraphics[width=14 cm]{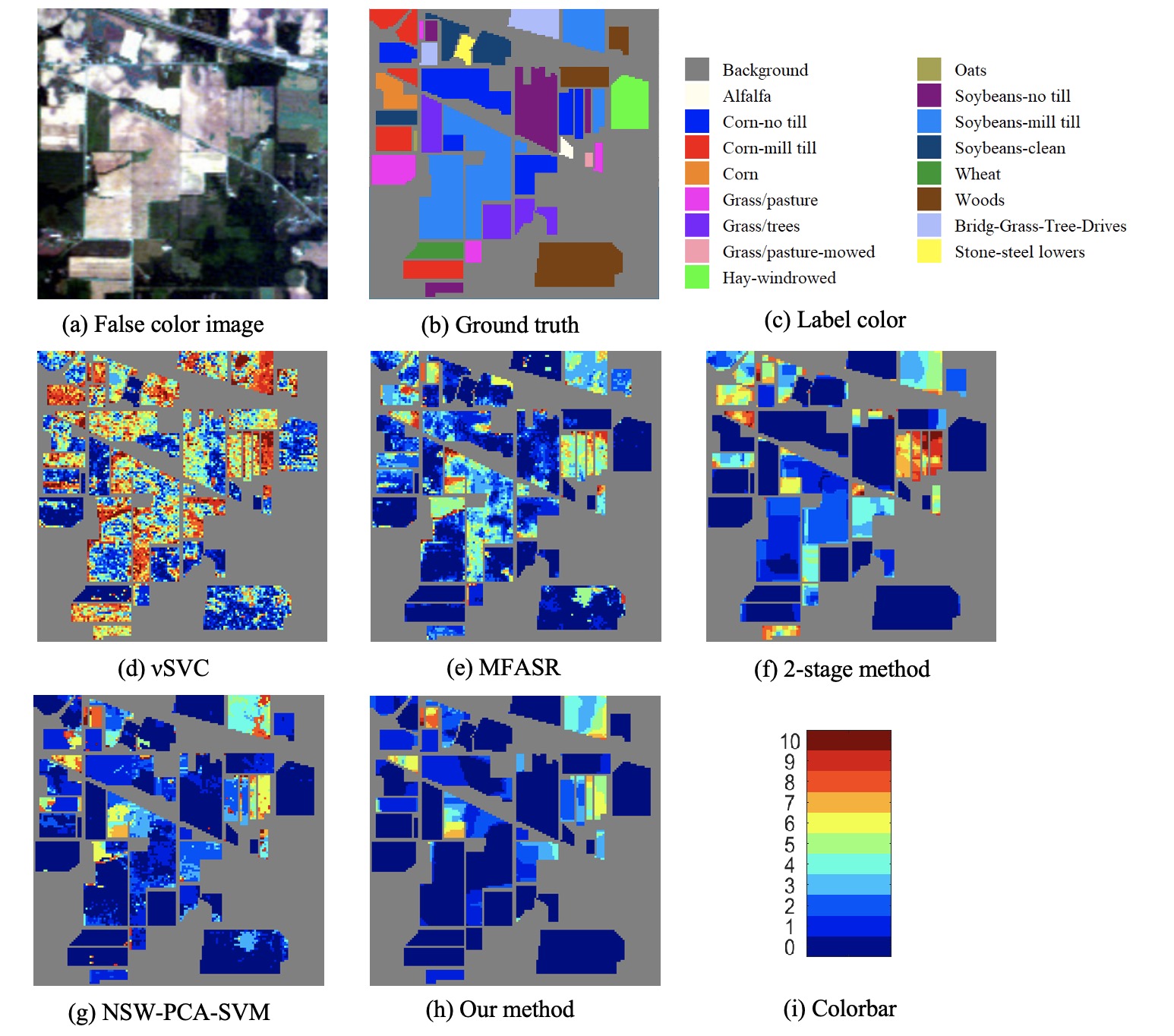}
\caption{Results for the Indian Pines data set. (\textbf{a}) The false color image. (\textbf{b})--(\textbf{c}) The ground truth and the corresponding label colors. (\textbf{d})--(\textbf{h}) The misclassification counts of different methods. (\textbf{i}) The colorbar representing the misclassification counts.
\label{IndianPines_result}}
\end{figure}

Figure~\ref{IndianPines_result} shows the ground-truth and error maps of misclassifications for the Indian Pines data set. Among them, $\nu$SVC, which uses only spectral information, produces the largest {portion} of misclassification and almost all classes have serious misclassification. The 2-stage method has poor classification results in the upper-left, upper-right and bottom regions, and the corresponding materials of these regions are Corn-mill till, Corn-no till, Soybeans-mill till and Soybeans-no till respectively. These classes have similar spectra, and the 2-stage method cannot distinguish them very well. MFASR method has a similar degree of misclassification as the 2-stage method. We see from Figure~\ref{IndianPines_result}(h) that our method, with the pre-processing stage, produces the best result {because it enhances the consistency of adjacent pixels, especially those pixels located in a large homogeneous area with similar interclass spectra}. Finally, when compared with NSW-PCA-SVM method, our method improves the result in most areas, especially for the Soybeans-mill till class. This shows that the smoothing TV step is very effective in {enforcing spatial connectivity} to increase the accuracy of the classification.

Table~\ref{Salinas} shows the {average classification results over 10 trials on the Salinas data set using 10 random pixels per class for training in each trial.} Our method also achieves the best performance in OA, AA and kappa when compared with the other four methods with a gain of at least $1.32\%$ in the accuracies. For Grapes-untrained class and Vinyard-untrained class, $\nu$SVC yields less than 60\% accuracy, indicating that the spectra of these two classes cannot provide enough information for {discrimination}. In comparison, the accuracies of our method for these two classes are enhanced a lot, nearly 40\%.

\begin{figure}[H]
\includegraphics[width=12 cm]{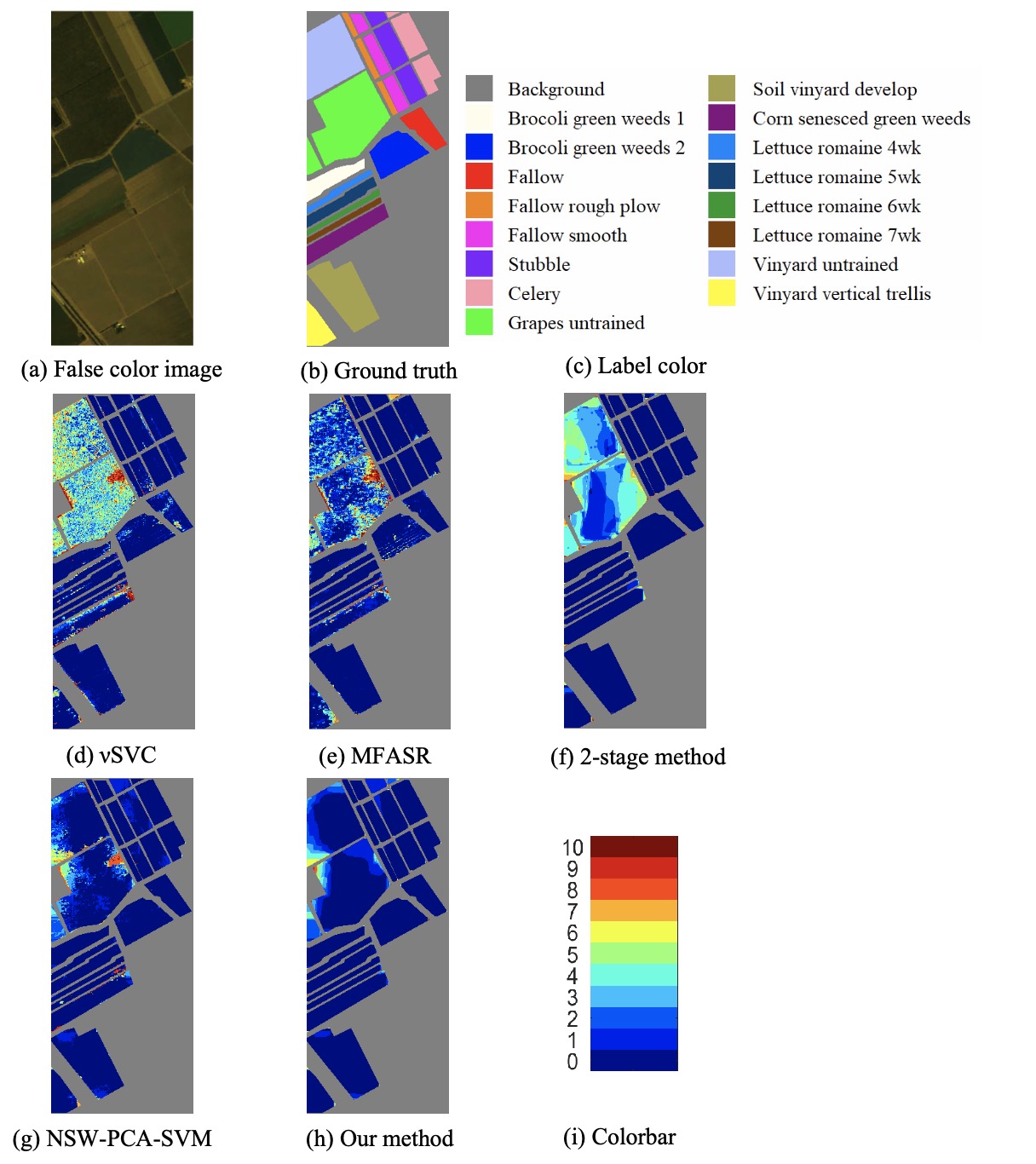}
\caption{Results for the Salinas data set. (\textbf{a}) The false color image. (\textbf{b})--(\textbf{c}) The ground truth and corresponding label colors. (\textbf{d})--(\textbf{h}) The misclassification counts of different methods. (\textbf{i}) The colorbar representing the misclassification counts.
\label{Salinas_result}}
\end{figure}

Figure \ref{Salinas_result} shows the ground-truth and error maps of misclassifications for the Salinas data set.
In Figure~\ref{Salinas_result}(d)--(f), we see that the $\nu$SVC, MFSAR and the 2-stage method all have large areas of misclassification in the Salinas data set. NSW-PCA-SVM method (Figure~\ref{Salinas_result}(g)) has a great improvement over the first three methods due to the pre-processing step, {but there is still a serious misclassification in the upper right corner of Grapes-untrained class}. Our method solves this problem by adding the denoising step to enhance {local} spatial homogeneity, see Figure~\ref{Salinas_result}(h). As a whole, the results show that the pre-processing and post-processing stages have a great effect on those classes with large homogeneous regions and insufficient spectral information.

\begin{table}[H]
\centering
\caption{Average classification accuracies over 10 trials for the Salinas data set with 10 random training pixels for each class.}
\label{Salinas}
\resizebox{0.8\textwidth}{!}{%
\begin{tabular}{cccccc}
\hline
\textbf{Class}	& \textbf{$\nu$SVC}	& \textbf{MFASR} & \textbf{2-stage method} & \textbf{NSW-PCA-SVM} & \textbf{Our method}\\
\midrule
Brocoli-green-weeds-1	&98.02\%&98.50\%&99.84\%&\textbf{99.86\%}&99.76\%\\
Brocoli-green-weeds-2 &97.70\%&95.85\%&99.78\%&\textbf{99.82\%}&99.79\%\\
Fallow &92.84\%&99.04\%&99.35\%&\textbf{99.92\%}&99.70\%\\
Fallow-rough-plow &98.64\%&99.68\%&98.17\%&\textbf{99.92\%}&98.95\%\\
Fallow-smooth &95.57\%&98.86\%&99.00\%&98.80\%&\textbf{99.36\%}\\
Stubble	&97.90\%&\textbf{99.72\%}&99.32\%&96.89\%&97.57\%\\
Celery	&98.74\%&96.37\%&99.12\%&\textbf{99.71\%}&98.93\%\\
Grapes-untrained &55.77\%&72.26\%&70.26\%&88.95\%&\textbf{93.72\%}\\
Soil-vinyard-develop &97.35\%&99.37\%&\textbf{99.78\%}&98.80\%&99.40\%\\
Corn-senesced-green-weeds  &79.17\%&89.31\%&98.54\%&95.77\%&\textbf{99.54\%}\\
Lettuce-romaine-4wk &92.02\%&98.15\%&99.36\%&99.40\%&\textbf{99.74\%}\\
Lettuce-romaine-5wk &97.52\%&99.48\%&99.73\%&\textbf{99.79\%}&99.71\%\\
Lettuce-romaine-6wk &98.18\%&97.23\%&99.64\%&97.70\%&\textbf{99.77\%}\\
Lettuce-romaine-7wk &89.58\%&92.53\%&97.78\%&92.59\%&\textbf{99.61\%}\\
Vinyard-untrained	&57.49\%&78.95\%&64.19\%&89.79\%&\textbf{92.20\%}\\
Vinyard-vertical-trellis &93.71\%&89.93\%&97.79\%&98.12\%&\textbf{99.22\%}\\
\hline
OA	&81.82\%&89.38\%&88.47\%&95.33\%&\textbf{97.15\%}\\
AA	&90.01\%&94.08\%&95.10\%&97.24\%&\textbf{98.56\%}\\
kappa &79.85\%&88.20\%&87.18\%&94.81\%&\textbf{96.82\%}\\
\hline
\end{tabular}%
}
\end{table}

\begin{figure}[H]
\includegraphics[width=12cm]{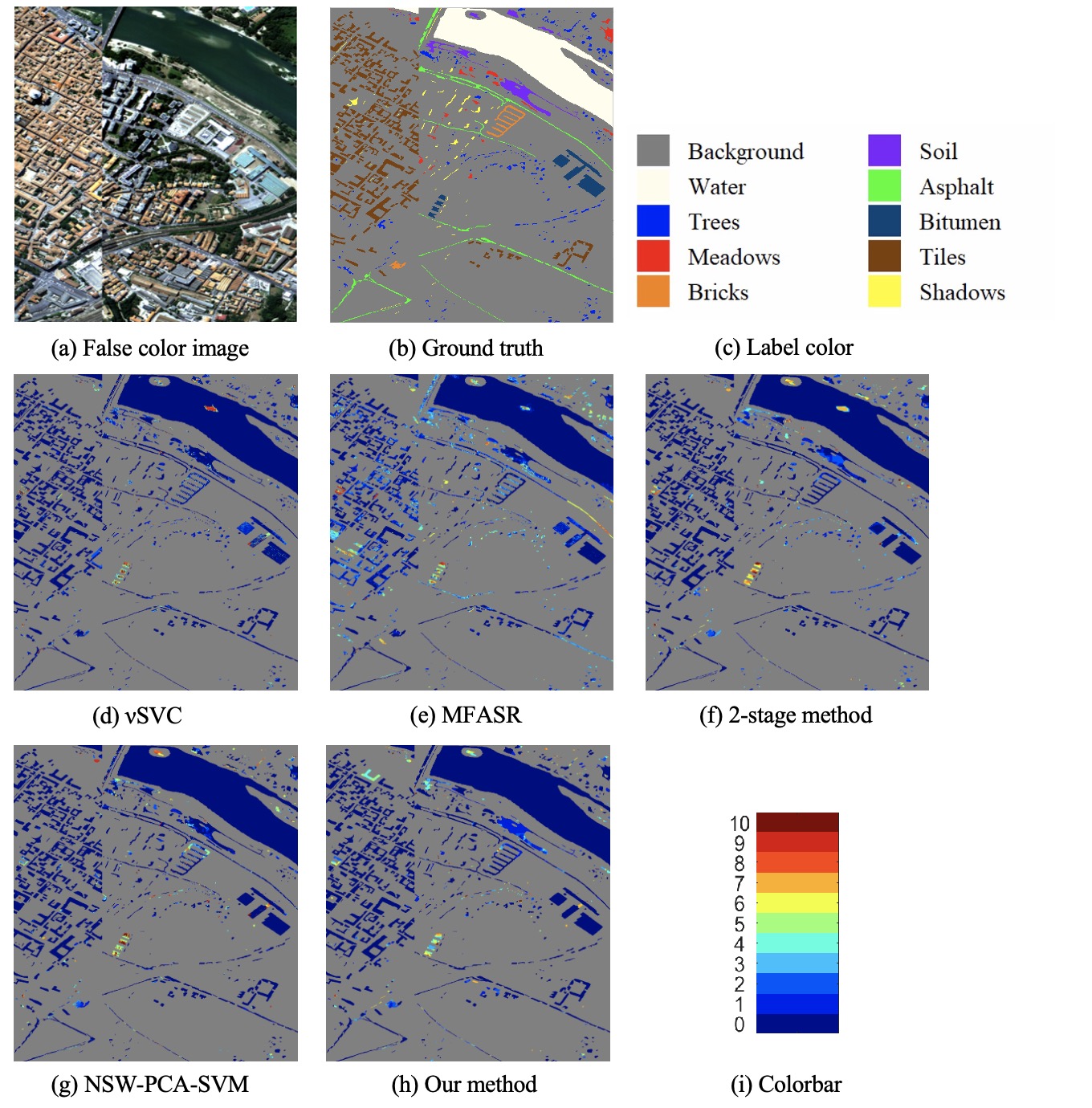}
\caption{Results for the Pavia Center data set. (\textbf{a}) The false color image. (\textbf{b})--(\textbf{c}) The ground truth and corresponding label colors. (\textbf{d})--(\textbf{h}) The misclassification counts of different methods. (\textbf{i}) The colorbar representing the misclassification counts.
\label{paviacenter_result}}
\end{figure}

{Table~\ref{paviacenter} shows the average classification results of Pavia Center data set over 10 trials with 10 random labeled pixels per class in each trial.} It consists of more small regions and slender categories, see Figure~\ref{paviacenter_result}(b). Our method is also the best one in all OA, AA and kappa coefficient. For water class, which is the largest region in the data set, our method achieves 100\% accuracy. While for those classes which do not have greatly ample spectral information, like Meadows class and Bitumen class, our method earns a gain of at least 3.5\% in accuracies.
In Figure~\ref{paviacenter_result}, which shows the misclassification map of the Pavia Center dataset, we see that $\nu$SVC has distinct misclassification in the middle of water class. Obviously, MFASR method has a worse result in almost all categories, especially for Trees and Tiles classes where $\nu$SVC has great classification results only using spectral information.
The 2-stage method and NSW-PCA-SVM method both have higher degree of misclassification in Bitumen class, mainly in the middle part of the image. Our method smooths the result, particularly for water class and Bitumen class in the middle of the image, which again shows the strength of the pre-processing step and post-processing step.

\begin{table}[t]
\centering
\caption{{Average classification accuracies over 10 trials } for the Pavia Center data set with 10 random training pixels for each class.}
\label{paviacenter}
\resizebox{0.7\textwidth}{!}{%
\begin{tabular}{cccccc}
\hline
\textbf{Class}	& \textbf{$\nu$SVC}	& \textbf{MFASR} & \textbf{2-stage method} & \textbf{NSW-PCA-SVM} & \textbf{Our method}\\
\midrule
Water	&99.02\%&99.60\%&99.56\%&\textbf{100\%}&\textbf{100\%}\\
Trees &81.58\%&73.06\%&76.29\%&85.99\%&\textbf{89.07\%}\\
Meadows &80.78\%&77.34\%&88.21\%&89.64\%&\textbf{93.14\%}\\
Bricks &75.65\%&91.99\%&\textbf{92.70\%}&81.42\%&92.05\%\\
Soil &78.80\%&86.20\%&84.58\%&\textbf{89.90\%}&86.26\%\\
Asphalt	&89.26\%&82.75\%&97.70\%&93.40\%&\textbf{97.79\%}\\
Bitumen	&80.64\%&88.02\%&87.64\%&88.30\%&\textbf{92.05\%}\\
Tiles &95.33\%&89.94\%&\textbf{99.18\%}&99.15\%&98.67\%\\
Shadows &\textbf{99.74\%}&96.75\%&99.30\%&99.27\%&97.84\%\\
\hline
OA	&93.86\%&92.58\%&96.53\%&97.04\%&\textbf{97.59\%}\\
AA	&86.76\%&87.29\%&91.68\%&91.90\%&\textbf{94.10\%}\\
kappa &91.37\%&89.61\%&95.09\%&95.80\%&\textbf{96.59\%}\\
\hline
\end{tabular}%
}
\end{table}

\begin{figure}[ht]
\centering

\subfigure[OA on the Indian Pines data set]{
\includegraphics[width=5cm,height=4cm]{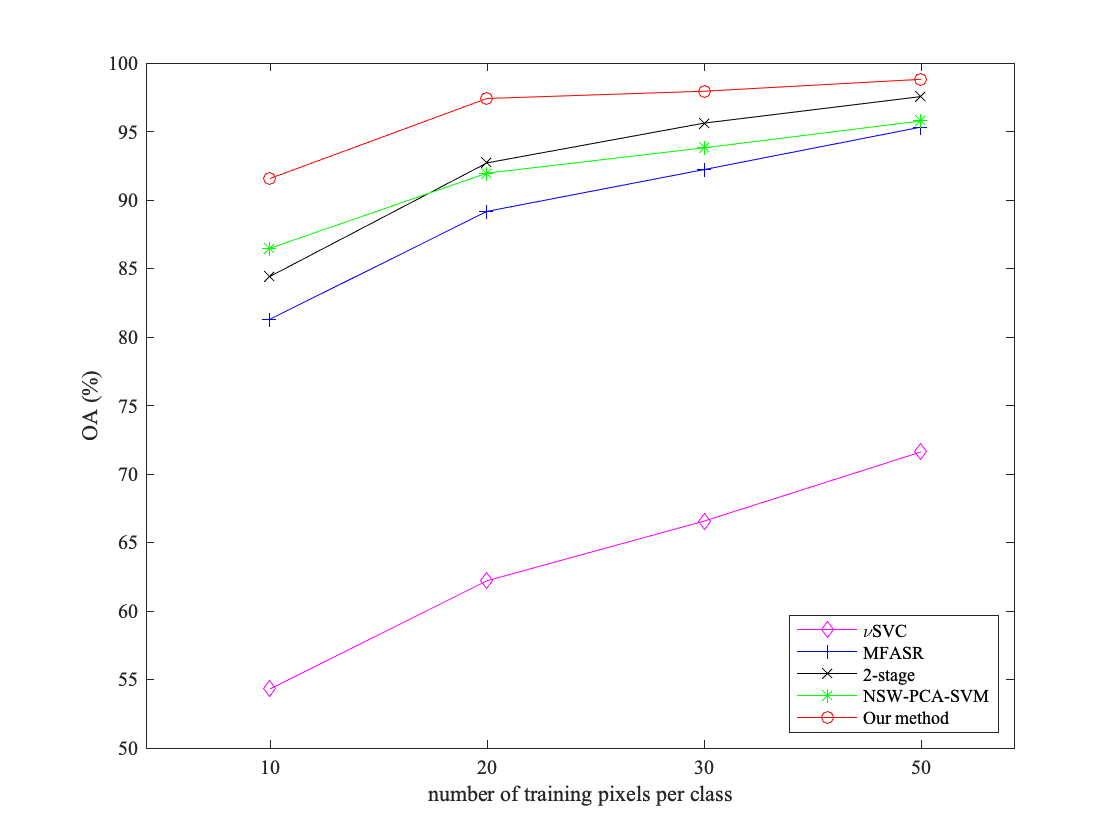}}
\subfigure[OA on the Salinas data set]{
\includegraphics[width=5cm,height=4cm]{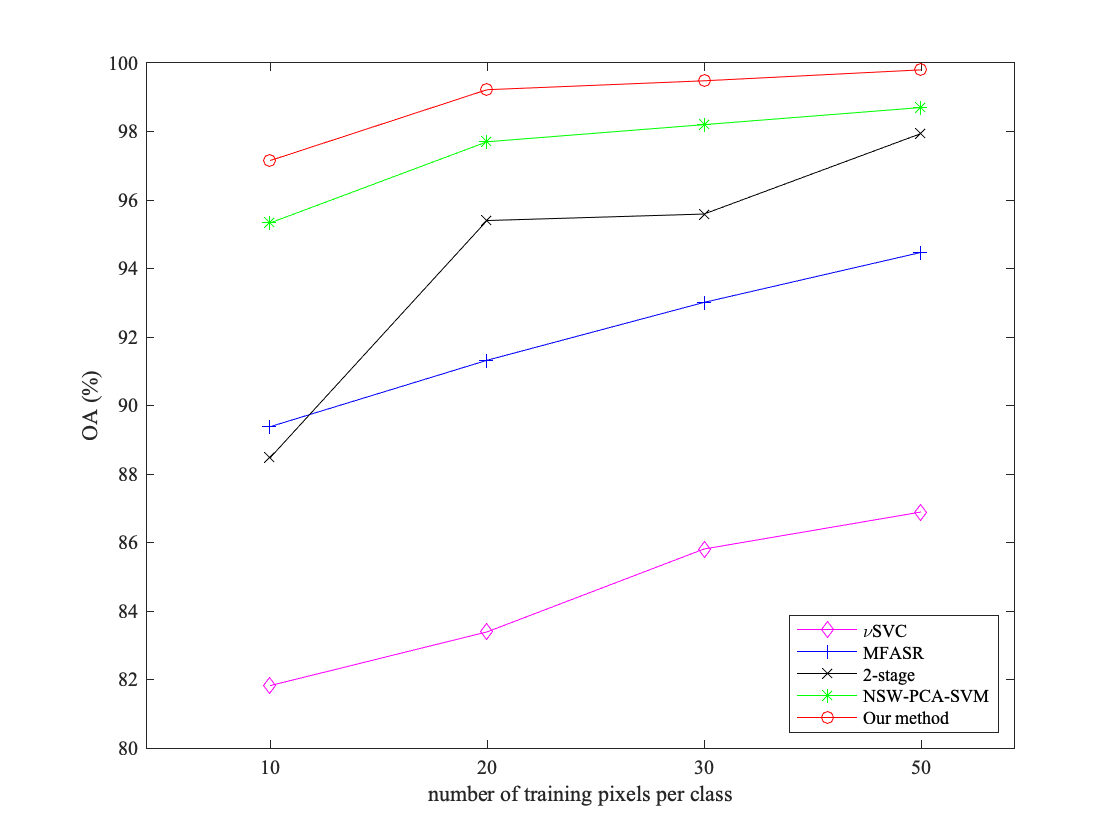}}
\subfigure[OA on the Pavia Center data set]{
\includegraphics[width=5cm,height=4cm]{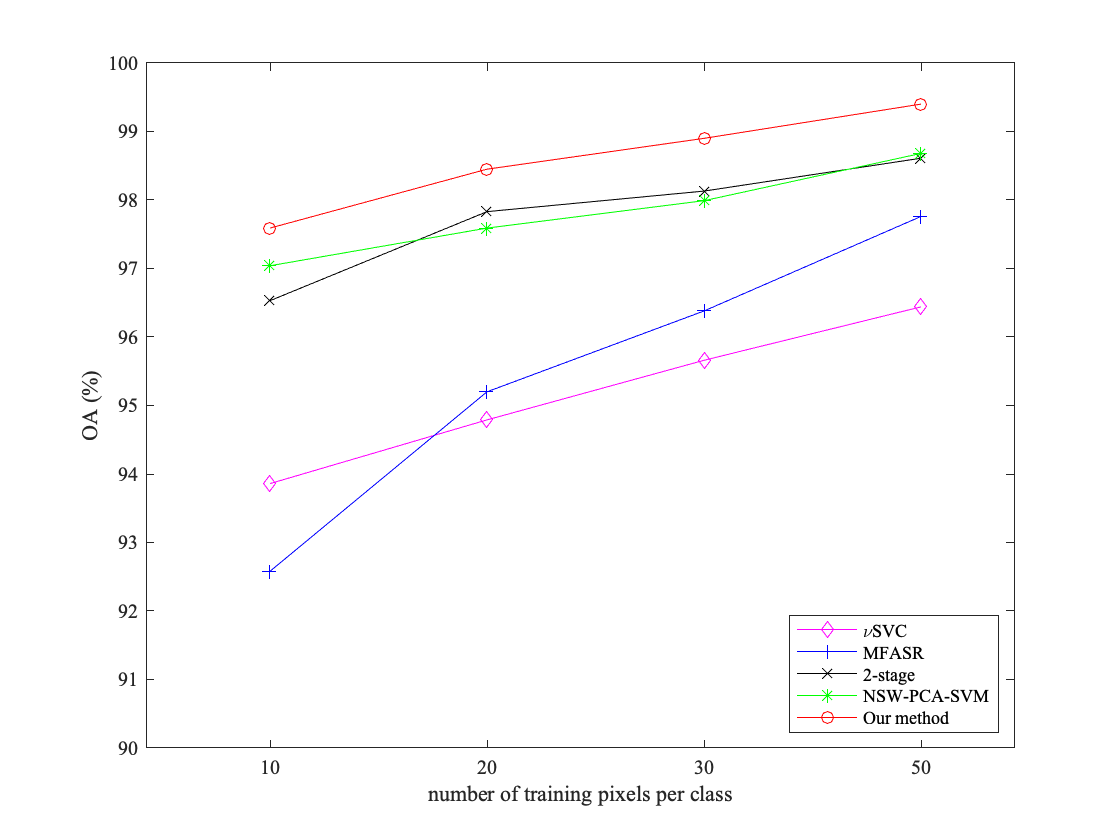}}

\subfigure[OA on the KSC data set]{
\includegraphics[width=5cm,height=4cm]{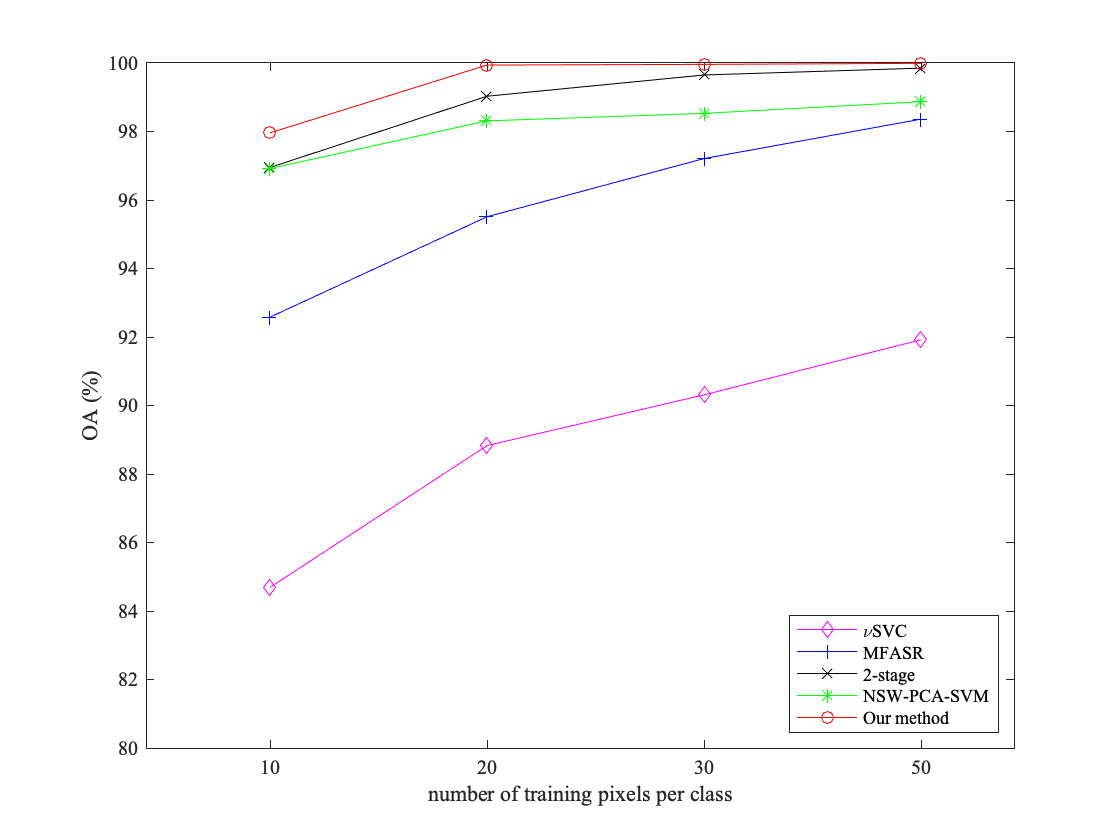}}
\subfigure[OA on the Botswana data set]{
\includegraphics[width=5cm,height=4cm]{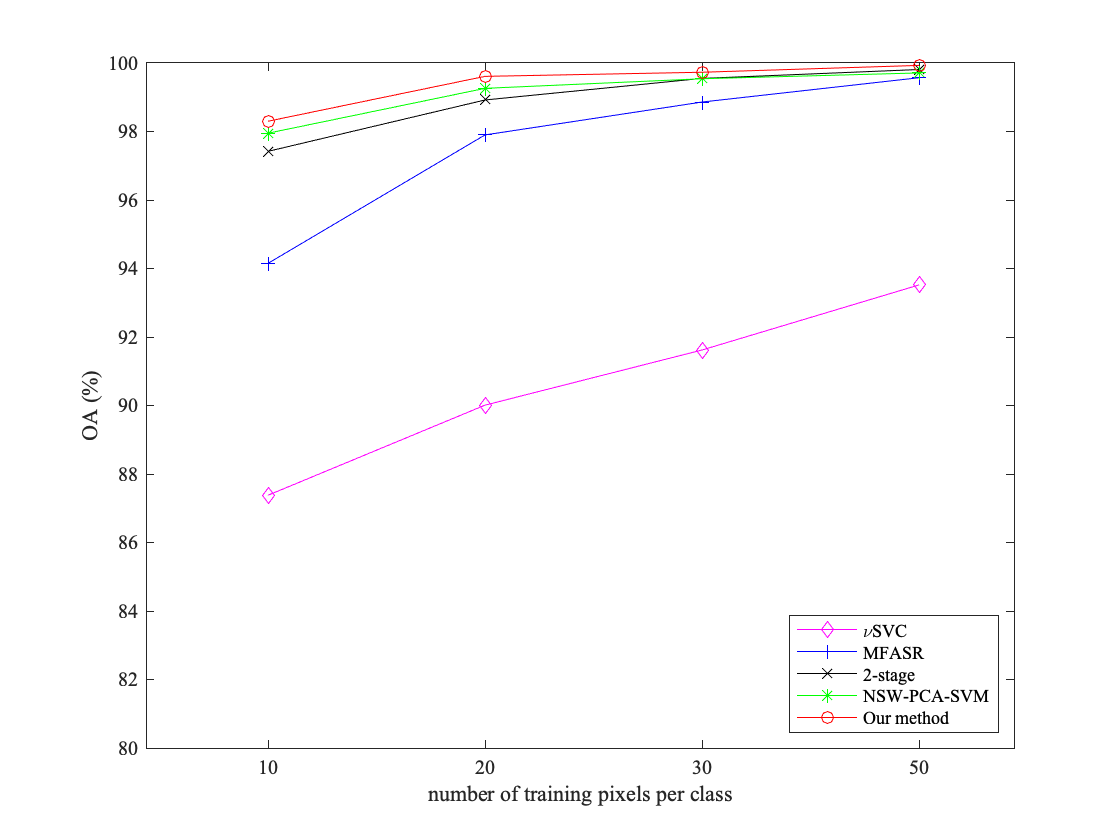}}
\subfigure[OA on the PaviaU data set]{
\includegraphics[width=5cm,height=4cm]{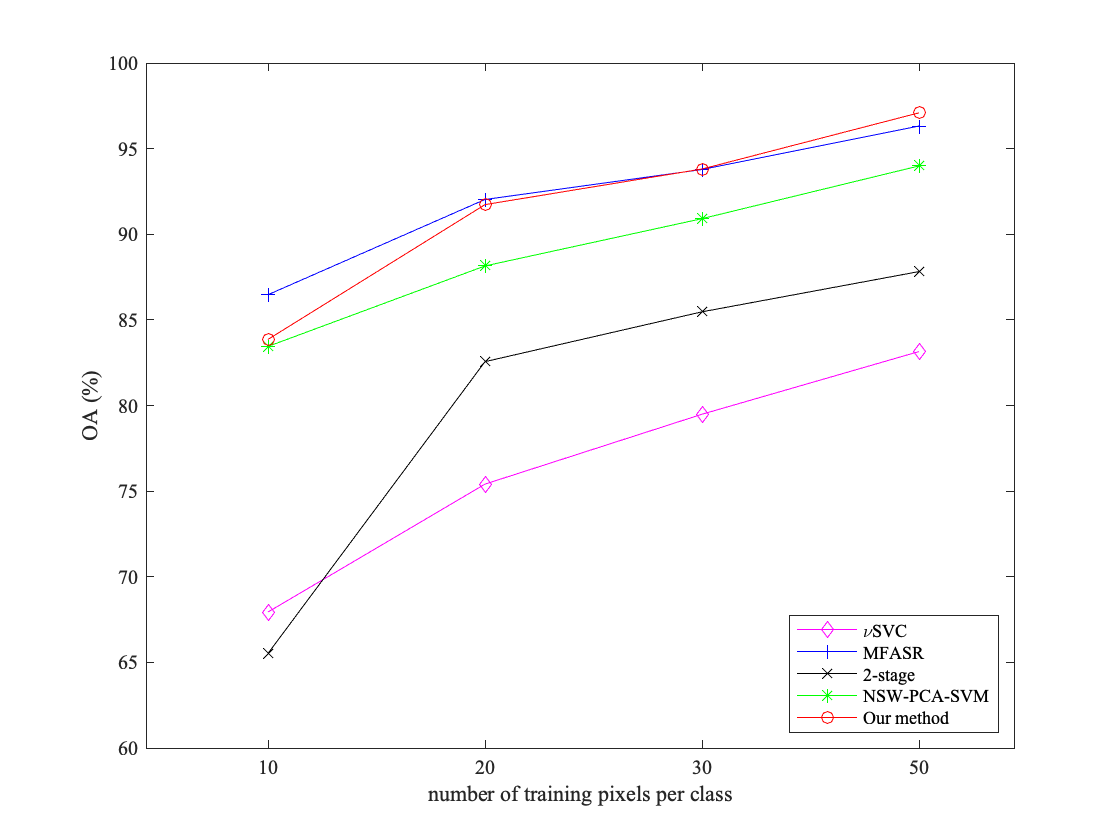}}

\caption{OAs ($y$-axis) for different data sets with different number of training pixels ($x$-axis).\label{OA}}
\end{figure}

{Figure~\ref{OA} show the overall averages (OAs) of different methods on the six data sets with different numbers of training pixels. For the first five data sets, our method achieves the best performance for all cases.
Pavia Center, KSC and Botswana data sets have more effective spectral information since $\nu$SVC already reaches more than 80\% accuracy. Our method is still enhanced a lot after adding steps for spatial information extraction, reaching more than 99\% accuracy with the increment of the training pixels.
One can see that the gain of accuracy of our method over the other methods increases when the number of labeled pixels decreases. This shows the advantage of our method as getting labeled data is always the most difficult part of any HSI classification problem.}
  
{Our method comes second only in one situation: the PaviaU data set with 10 training pixels per class. However, with the increment of training pixels, our method gradually exceeds MFASR. We note that MFASR generally fares only better than $\nu$SVC in the other five data sets.} From Figures \ref{OA}(b) and (f), we see that the 2-stage method can be bad if the number of pixels is small, but this deficiency can be resolved by our pre-processing stage as it enhances the consistency of the neighboring pixels so that we only need a smaller training set.

\begin{figure}[ht]
\includegraphics[width=12 cm]{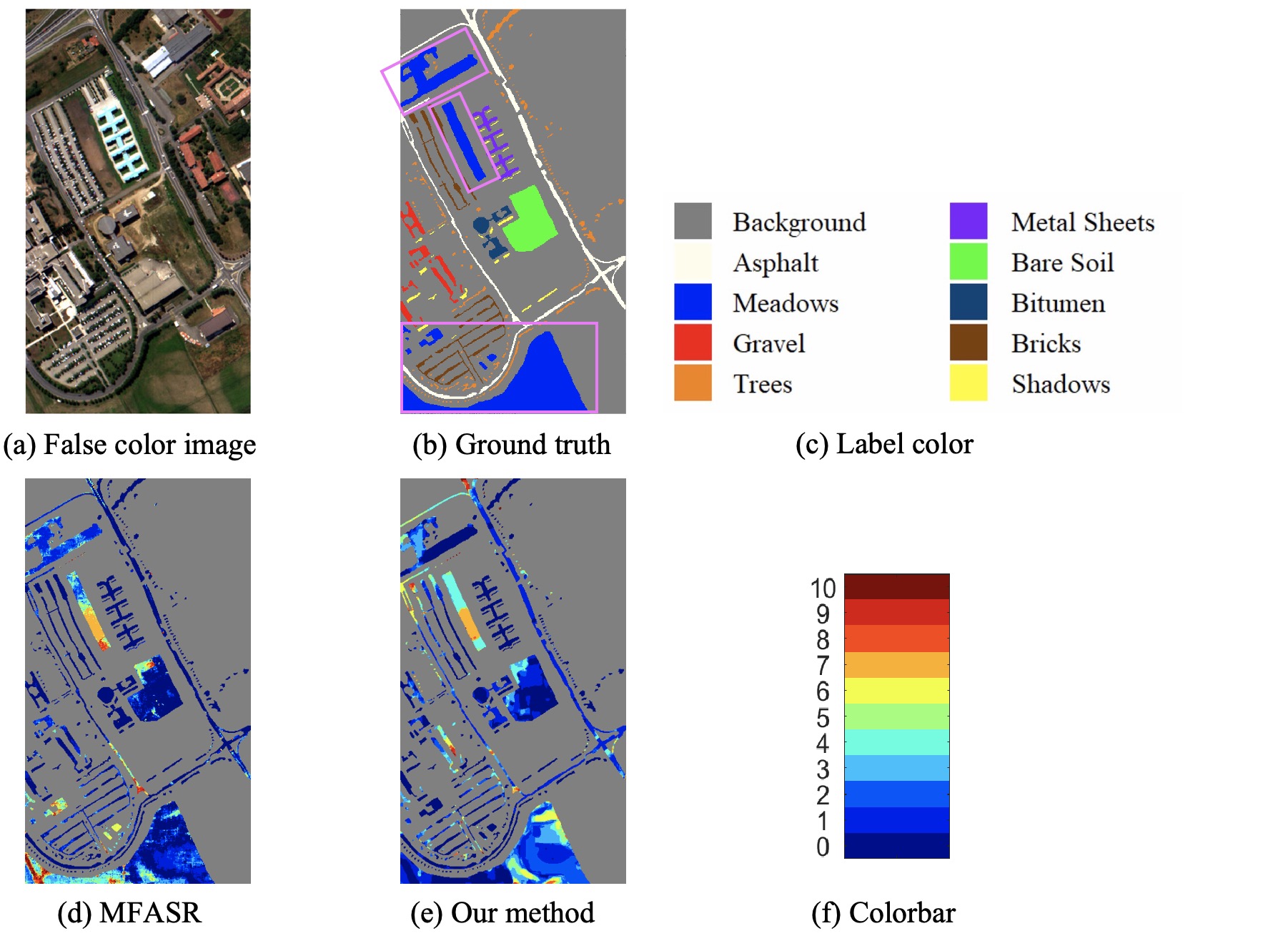}
\caption{{Results for the PaviaU data set. (\textbf{a}) The false color image. (\textbf{b})--(\textbf{c}) The ground truth and the corresponding label colors. (\textbf{d})--(\textbf{e}) The misclassification counts of MFASR and our method. (\textbf{f}) The colorbar representing the misclassification counts.
\label{PaviaU_result}}}
\end{figure}

\begin{figure}[H]
\includegraphics[width=13 cm]{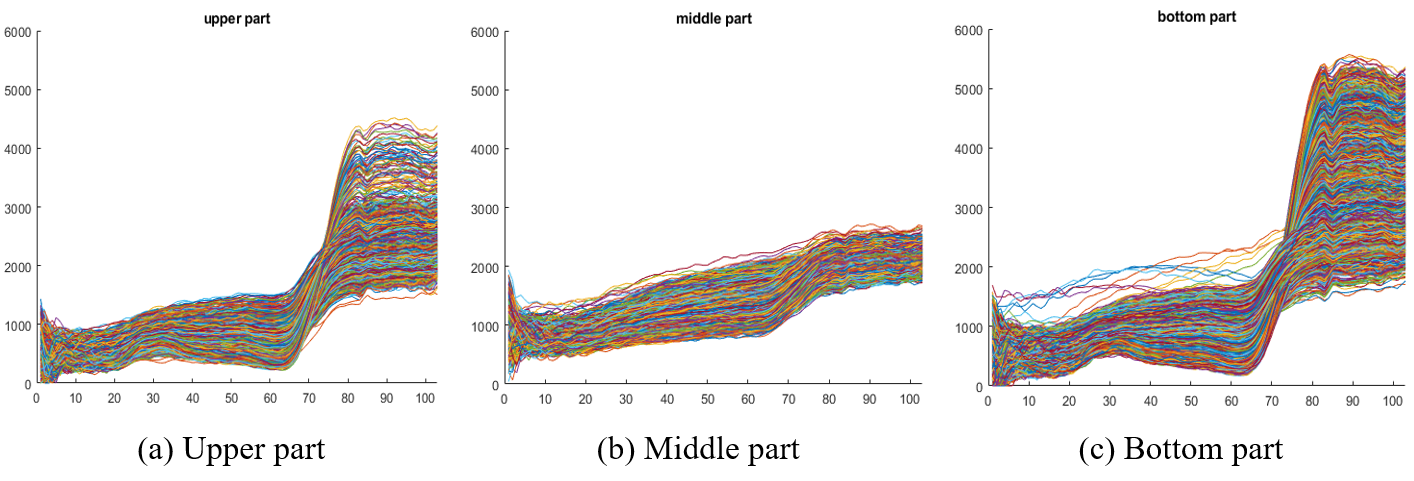}
\caption{The spectra of the Meadows class in the PaviaU data set: (\textbf{a}) In the upper part of the image. (\textbf{b}) In the middle part of the image. (\textbf{c}) In the bottom part of the image. They show that the spectra of the Meadows class in the middle part of the image vary significantly from the spectra of the Meadows class in other parts of the image. \label{3parts}}
\end{figure}

According to Figure~\ref{PaviaU_result}(b), in the PaviaU data set, the distribution of the pixels in the same category is relatively scattered, especially for the classes of Asphalt, Meadows, Bricks and Shadows. In addition, the shapes of many {regions} are slender and long where
MFASR seems to perform the best, see Figure~\ref{PaviaU_result}(d).
{We notice that no method has a good classification result for the Meadows class in the middle part of the image.} {Based on the ground-truth in Figure~\ref{PaviaU_result}(b), Meadows are in three different locations in the image: upper, middle, and lower parts as marked by the pink boxes in the figure. Their corresponding spectra are shown in Figure~\ref{3parts}.} From the figure we see that the spectra of the Meadow pixels in the middle part of the image vary greatly from the Meadow pixels in the other parts of the image, and this results in the difficulty in correctly classifying them.

\begin{figure}[H]
\centering

\subfigure[AA on the Indian Pines data set]{
\includegraphics[width=5cm,height=3.8cm]{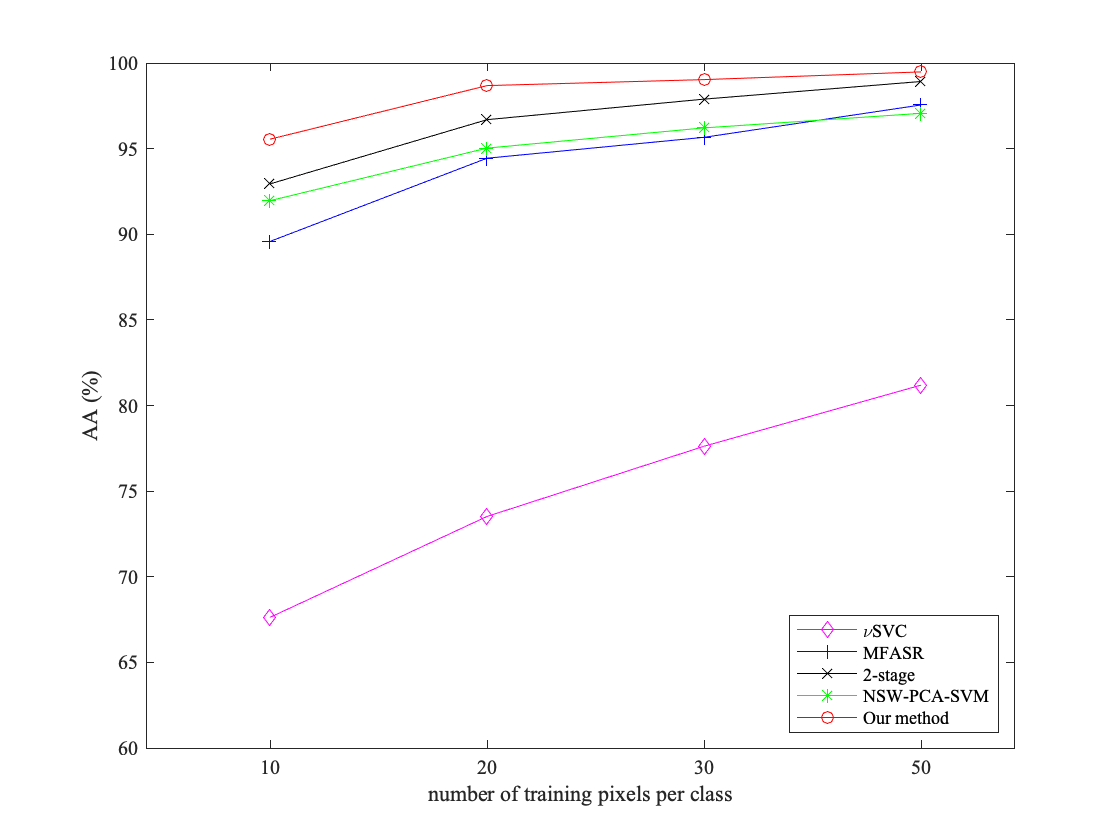}}
\subfigure[AA on the Salinas data set]{
\includegraphics[width=5cm,height=3.8cm]{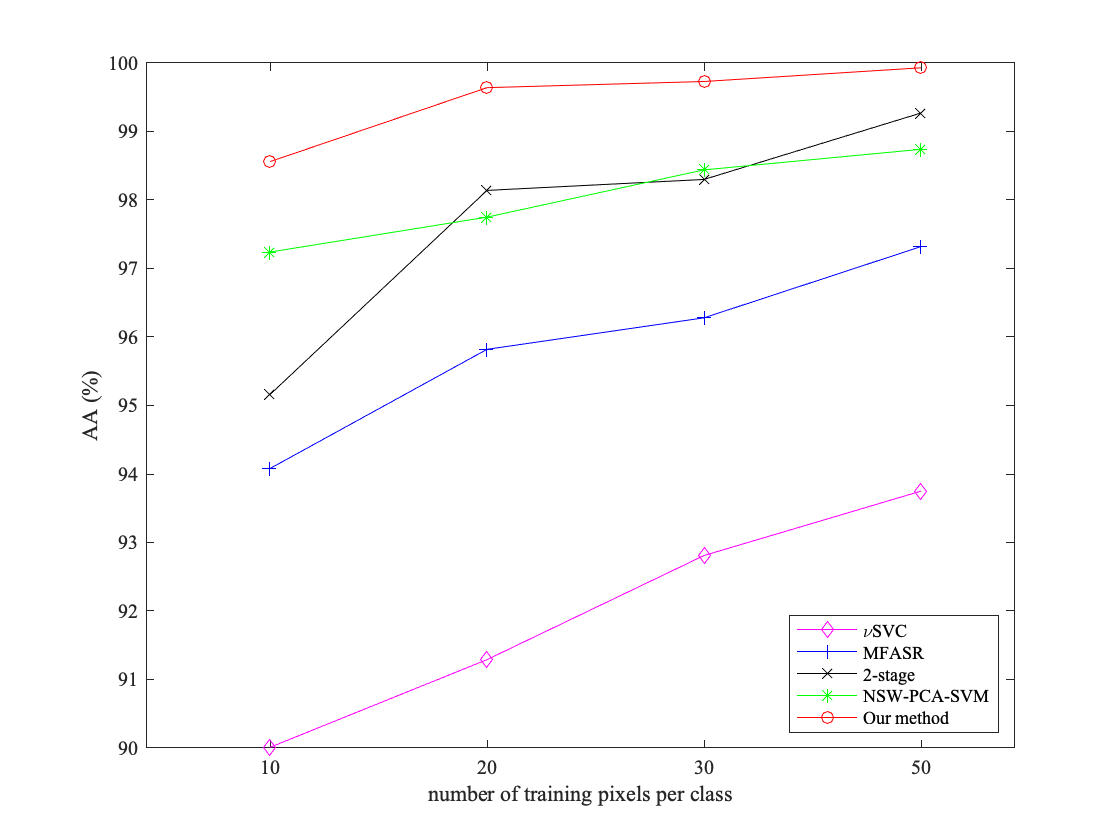}}
\subfigure[AA on the Pavia Center data set]{
\includegraphics[width=5cm,height=3.6cm]{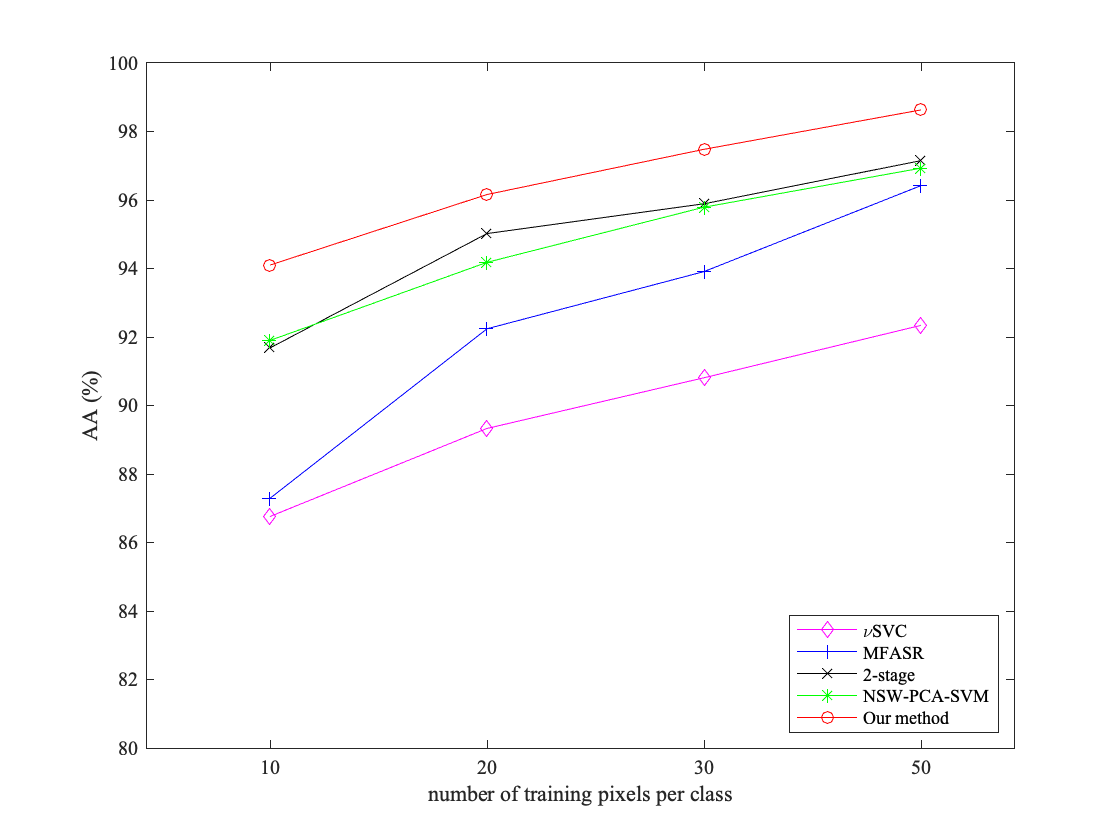}}

\subfigure[AA on the KSC data set]{
\includegraphics[width=5cm,height=3.8cm]{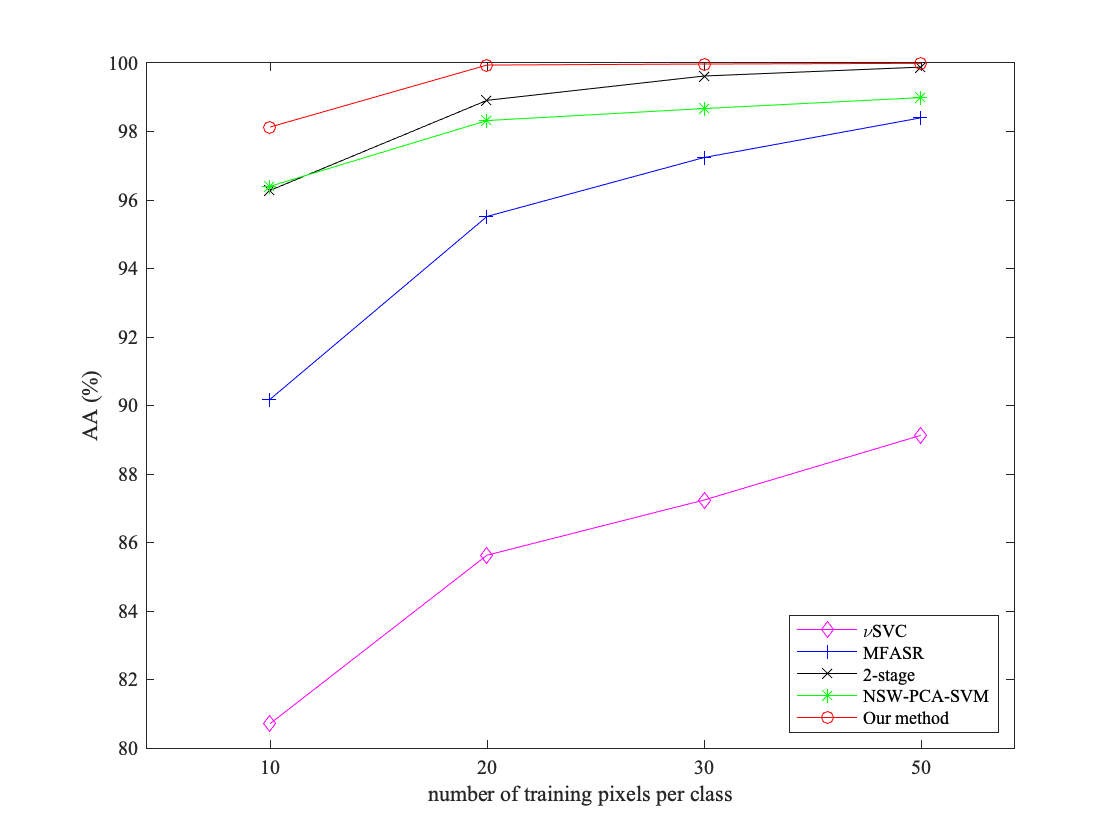}}
\subfigure[AA on the Botswana data set]{
\includegraphics[width=5cm,height=3.8cm]{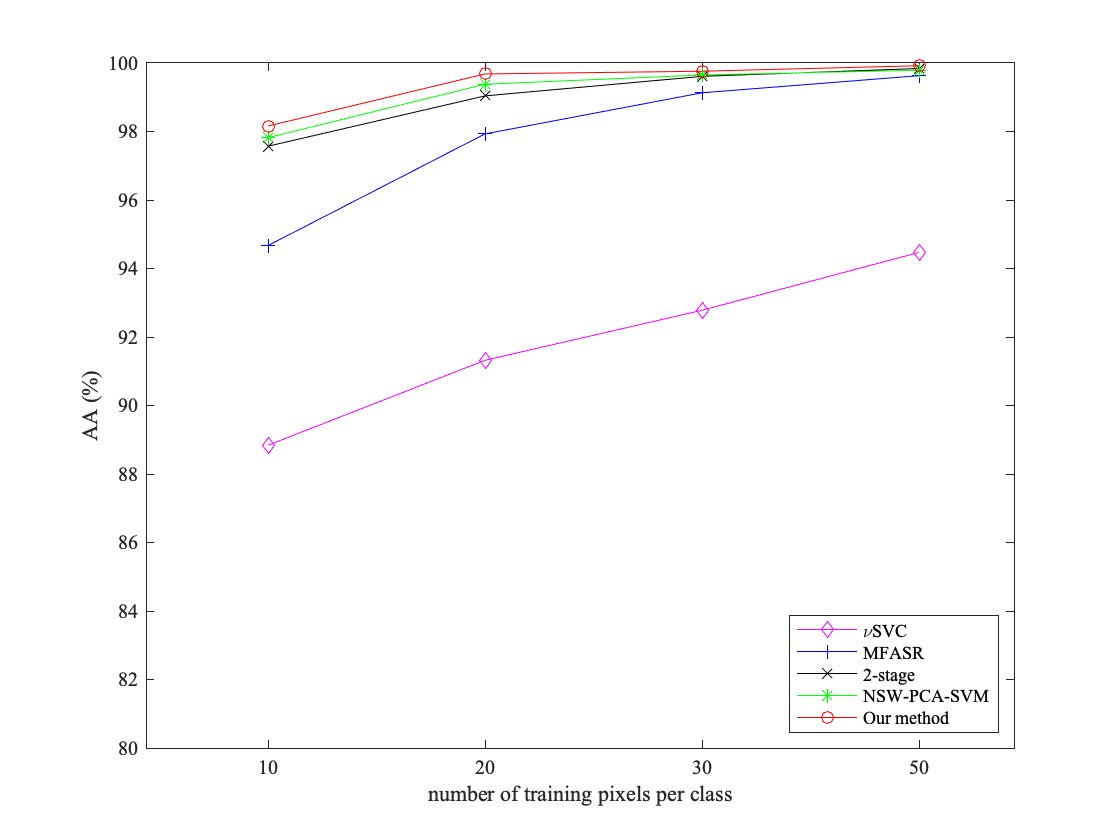}}
\subfigure[AA on the PaviaU data set]{
\includegraphics[width=5cm,height=3.8cm]{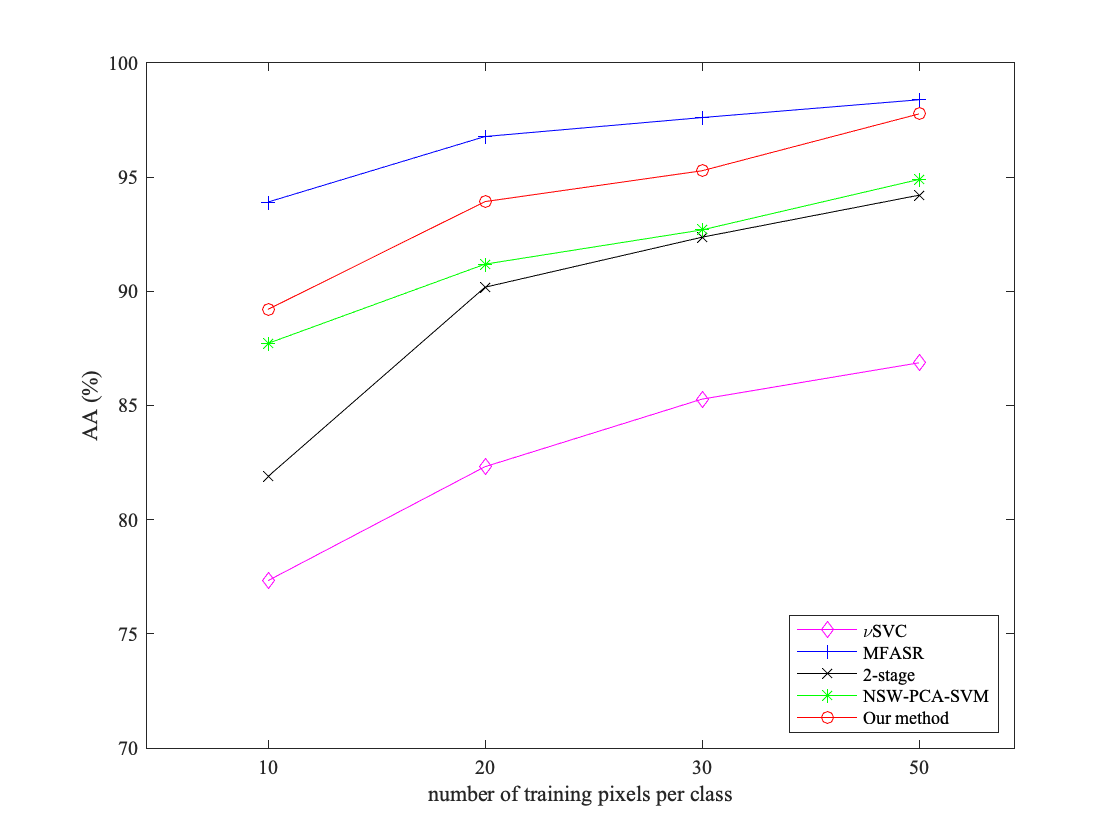}}

\caption{AAs ($y$-axis) for different data sets with different number of training pixels ($x$-axis).\label{AA}}
\end{figure}

Figure~\ref{AA} shows the average accuracy (AAs) of different methods on the six data sets with different numbers of training pixels. Our method still performs the best on the first five data sets, no matter how many labeled pixels are utilized. For Salinas, KSC and Botswana data set, the AAs are around 98\% even in the case that 10 labeled pixel are available and attain more than 99\% once there are more labeled pixel available for training. Only for the last data set, PaviaU data set, see Figure~\ref{AA}(f), our method attains the second highest accuracy. Nonetheless, with the increment of training pixels, the gain of our method over MFASR becomes smaller. 

\begin{figure}[H]
\centering

\subfigure[kappa on the Indian Pines data set]{
\includegraphics[width=5cm,height=3.8cm]{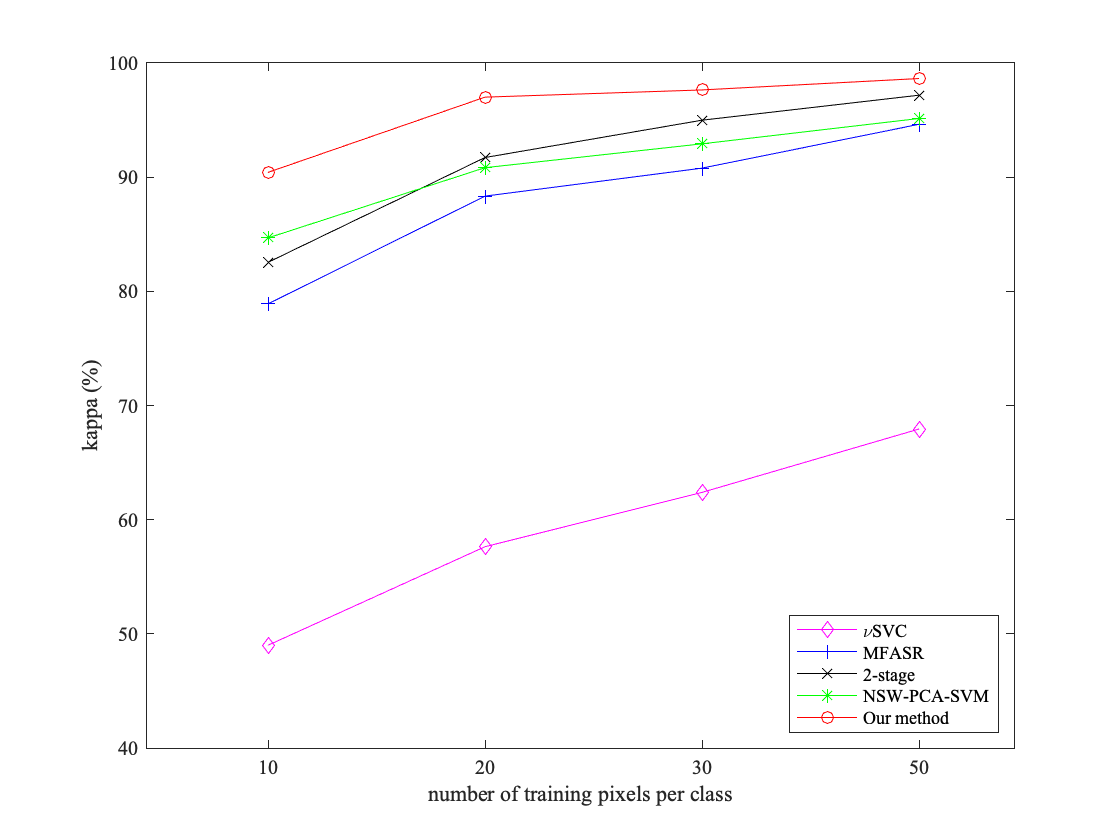}}
\subfigure[kappa on the Salinas data set]{
\includegraphics[width=5cm,height=3.8cm]{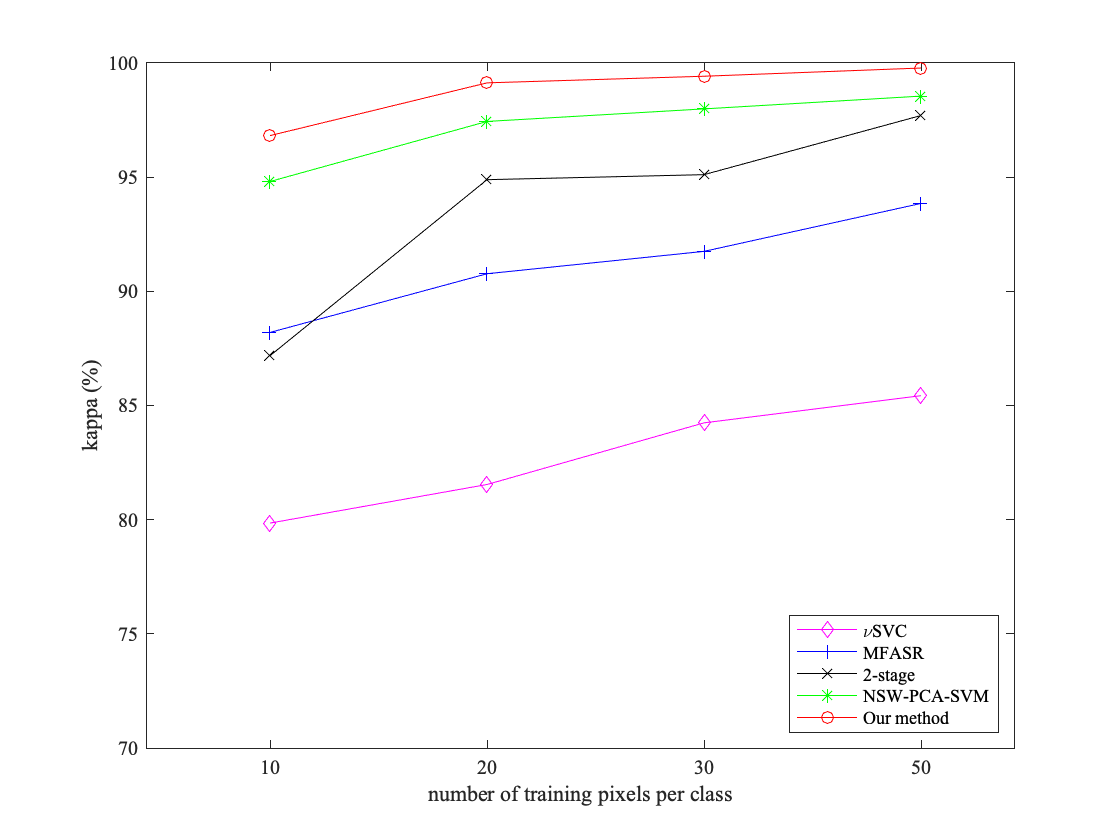}}
\subfigure[kappa on the Pavia Center data set]{
\includegraphics[width=5cm,height=3.8cm]{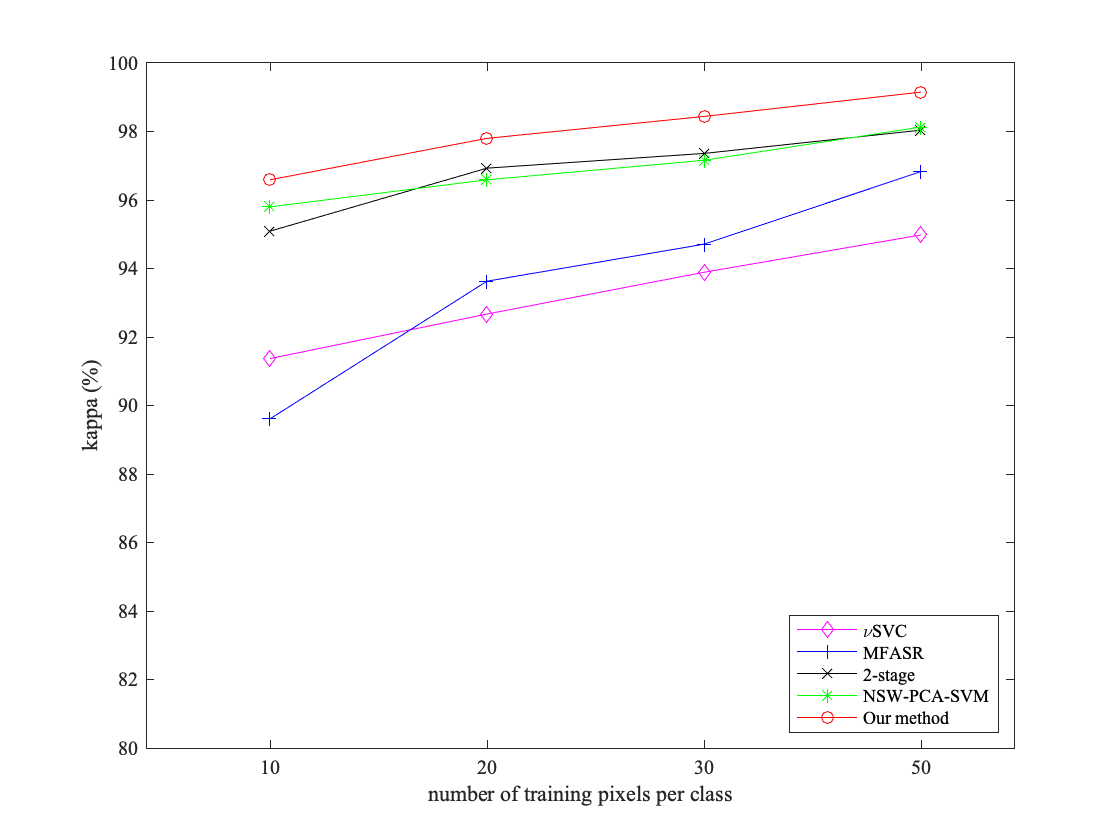}}

\subfigure[kappa on the KSC data set]{
\includegraphics[width=5cm,height=3.8cm]{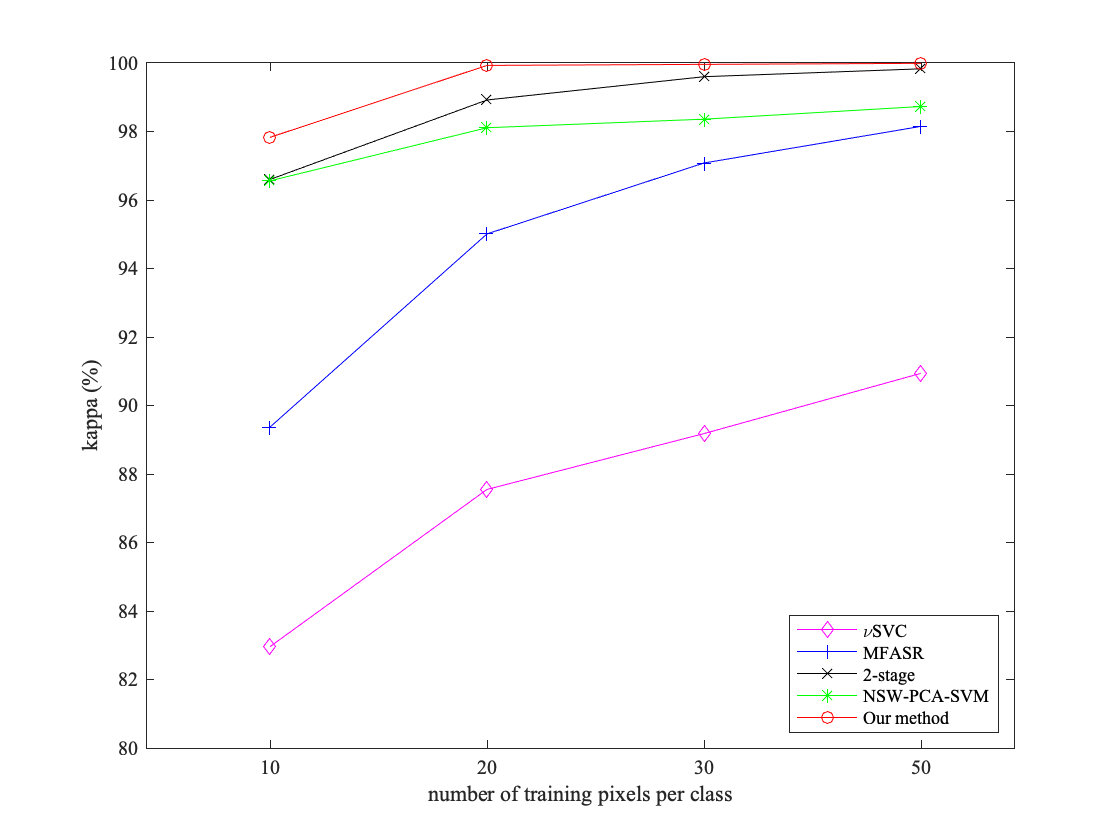}}
\subfigure[kappa on the Botswana data set]{
\includegraphics[width=5cm,height=3.8cm]{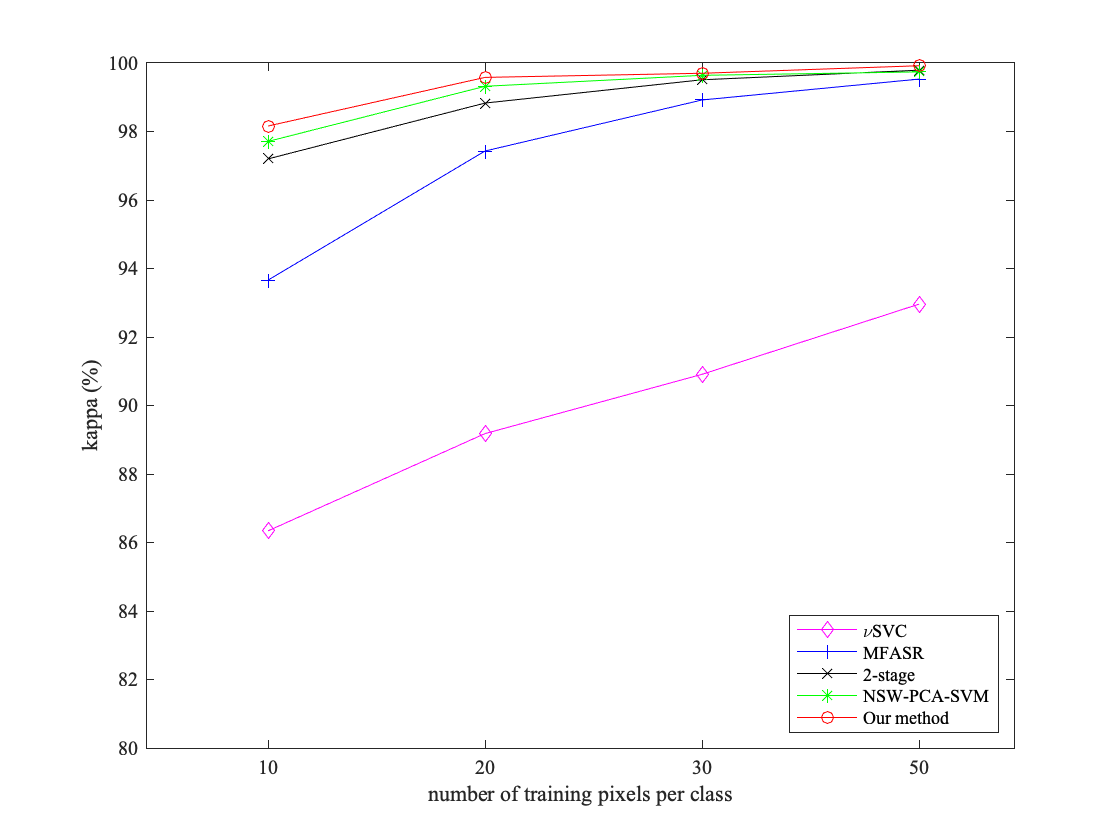}}
\subfigure[kappa on the PaviaU data set]{
\includegraphics[width=5cm,height=3.8cm]{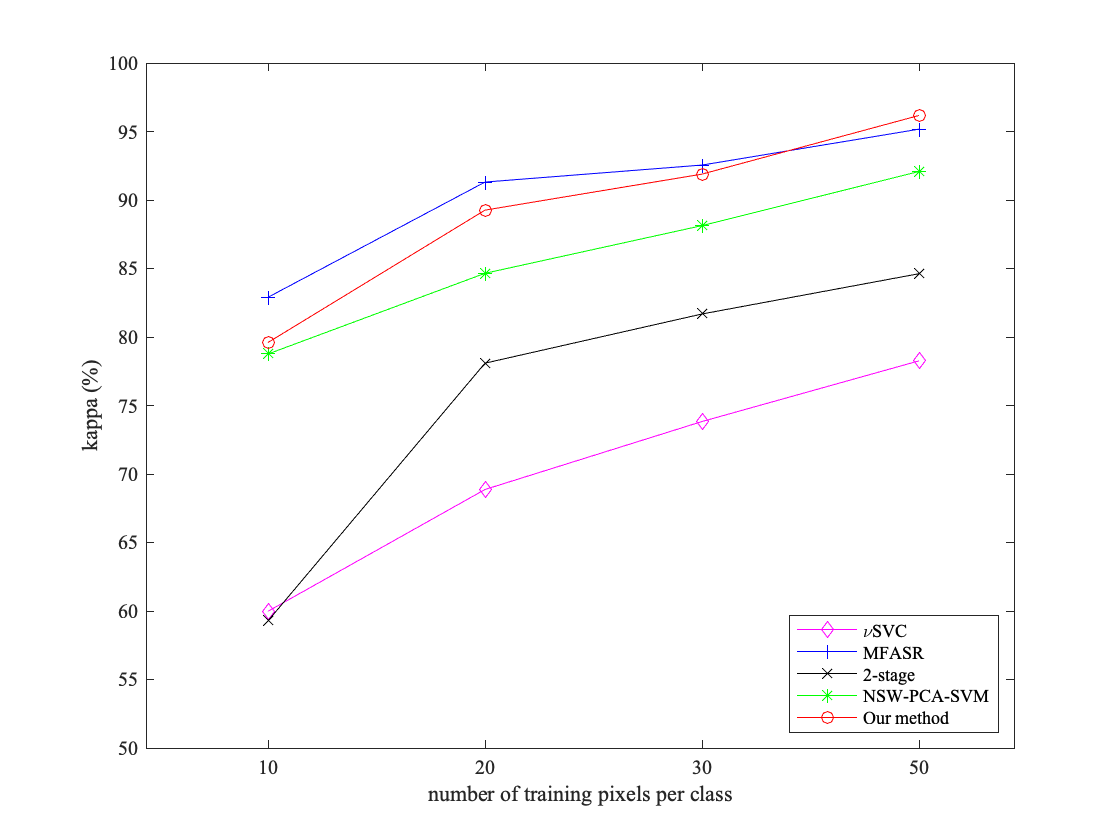}}

\caption{kappas ($y$-axis) for different data sets with different number of training pixels ($x$-axis).\label{kappa}}
\end{figure}

Figure~\ref{kappa} shows the kappas of different methods on the six data sets when different numbers of labeled pixels are used for training. Similar to the results of OAs and AAs, our method gives the highest accuracy on the first five data sets in all cases. For the PaviaU data set, our method is the second when there are small number of training pixels and outperforms MFASR when more training pixels are adopted.

To sum up, these figures clearly show the accuracy of our method over the other four methods on six data set in three different error metrics (OA, AA, and kappa) regardless of the number of training pixels used. We are only second to MFASR in PaviaU database. However, MFASR fares the worst for all the other five datasets except when compared to $\nu$SVC. The figures also show that the gain of our method over the other methods increases as the number of training pixels decrease, which attest to the importance of our method.

\unskip

\section{Discussions}

\subsection{Parameters for each method}
Table~\ref{para} shows the number of parameters for all methods mentioned in this paper.
In the experiments, the parameters are chosen as follows. For {$\nu$SVC} method and the first stage of the 2-stage method (which is also a $\nu$SVC method), there are two parameters and they are obtained by a 5-fold cross validation \cite{5fold}.
For the 2-stage method, the  three parameters in the second stage are chosen by trial-and-error such that it gives the highest classification result. For {MFASR} method, the ten {optimal}  parameters are chosen {as mentioned in \cite{Fang2017MFASR} by trial-and-error.}
For NSW-PCA-SVM method, the optimal window size and the optimal number of principal components are chosen by trial-and-error, while the parameters of {SVM} are chosen by a 5-fold cross validation.

For our method, there are 7 parameters in total. The window size and the number of principal components in the pre-processing stage are chosen by trial-and-error. The two parameters in $\nu$SVC are obtained automatically by a 5-fold validation. In the post-processing stage (see (\ref{eq4})), the regularization parameter $\beta_1$ is obtained by trial-and-error while $\beta_2$ can be fixed as 4, as the solution is robust against this parameter. When (\ref{eq4}) is solved by ADMM, there is a parameter governing the convergence rate and we set it always to 5. Thus in essence, there are only three parameters to be tuned by hand. Table~\ref{para_value} shows the values of these three parameters for the different data sets with 10 training pixels per class.

\begin{table}[htb]
\centering
\caption{The number of parameters in different methods.}
\label{para}
\resizebox{0.8\textwidth}{!}{%
\begin{tabular}{cccccc}
\hline
\textbf{} & \textbf{$\nu$SVC} & \textbf{MFASR} & \textbf{2-stage method} & \textbf{NSW-PCA-SVM} & \textbf{Our method}\\
\midrule
Number of parameters & 2 & 10 & 5 & 4 & 7\\
\hline
\end{tabular}
}
\end{table}
\unskip

\begin{table}[H]
\centering
\caption{The values of the parameters in our method for different data sets with 10 training pixels.}
\label{para_value}
\resizebox{0.6\textwidth}{!}{%
\begin{tabular}{cccc}
\hline
\textbf{} & size of window & principal component number& \textbf{$\beta_1$ }   \\
\midrule
\textbf{Indian Pines} & 21 & 25 & 0.2 \\
\textbf{Salinas} & 29 & 41 & 0.8 \\
\textbf{Pavia Center} & 11 & 9 & 0.2 \\
\textbf{KSC} & 17 & 67 & 0.1 \\
\textbf{Botswana} & 17 & 9 & 0.2 \\
\textbf{PaviaU} & 5 & 23 & 0.2 \\
\hline
\end{tabular}
}
\end{table}
\unskip

\subsection{Computation times for each method}

We test the computation times for all data sets with different methods, which only represent the running time of different algorithms and do not include the time needed to find the optimal parameters. {Table~\ref{time} shows the results in the case of 10 training pixels for each class.} {$\nu$SVC requires the least amount of time when compared with the other four methods since it does not need to pre-process or post-process the data. The 2-stage method needs a little more extra time compared with $\nu$SVC because of the denoising step. However, it has much higher accuracy than that of $\nu$SVC, see Figure~\ref{OA}.}
NSW-PCA-SVM method requires more time for reconstructing pixels and compressing the data. For our method, there is an additional denoising stage compared with NSW-PCA-SVM, thus we need a little more total computation time. However, the results of our method is enhanced a lot once we add the denoising stage, see Figures~\ref{OA}, \ref{AA} and \ref{kappa}. Basically, MFASR needs the longest time, which is because of the inner product between feature dictionaries and feature matrices. It takes nearly seven times as long as ours for PaviaU data set where it achieves the best performance. We emphasize that although our method is not the fastest (the fastest is $\nu$SVC), the accuracy of our method, especially for very small training data sets, can more than offset this drawback since the most time-consuming task in HSI classification is usually the labeling of the training pixels.

\begin{table}[htb]
\centering
\caption{Comparison of computation times (in seconds) for 10 training pixels.}
\label{time}
\resizebox{0.7\textwidth}{!}{%
\begin{tabular}{cccccc}
\hline
\textbf{Class}	& \textbf{$\nu$SVC}	& \textbf{MFASR} & \textbf{2-stage method} & \textbf{NSW-PCA-SVM} & \textbf{Our method}\\
\midrule
\textbf{Indian Pines} & 4.330 & 193.562 & 9.220 & 80.468&86.630\\
\textbf{Salinas} & 16.595 &985.179& 95.119&898.164&972.684\\
\textbf{Pavia Center} & 36.954 & 2197.981 & 255.212& 251.459&532.623\\
\textbf{KSC} & 2.152& 115.248 & 57.687 & 32.319&83.931\\
\textbf{Botswana} & 1.708 & 63.254 &65.990& 23.353&82.518\\
\textbf{PaviaU} & 5.045 & 590.348 & 58.547 & 42.361&92.824\\
\hline
\end{tabular}%
}
\end{table}

\section{Conclusion}\label{Conclusion}
{In this paper, we propose a new method which makes full use of the spatial and spectral information. Before classification, NSW and PCA are used to extract spatial information from the HSI and reconstruct the data. They enhance the {consistency} of the neighboring pixels so that we only need a smaller training set.  After that, $\nu$SVC is used to estimate the pixel-wise probability map of each class. Finally, a smoothed total variation model, which enhances spatial homogeneity {in the probability tensor}, is applied to classify the HSI into different classes. Compared with the other methods, our new method achieves the best overall accuracy, average accuracy, and kappa on six data sets except only for the PaviaU data set where we achieve the second best. The gain in accuracy of our method over the other methods increases when the number of training pixels available decreases. Our method is therefore of great practical significance since expert annotations are often expensive and difficult to collect.}

In the future, we will try to improve and develop new methods for adaptively selecting neighborhood pixels, which will be more helpful for those data sets which include more irregular regions like the PaviaU data set. In addition, different spatial filters will also be considered to extract spatial information and combine them with our method here.

\printbibliography 

\end{document}